\documentclass[letterpaper]{article} % DO NOT CHANGE THIS
\usepackage{aaai2026}  % DO NOT CHANGE THIS
\usepackage{times}  % DO NOT CHANGE THIS
\usepackage{helvet}  % DO NOT CHANGE THIS
\usepackage{courier}  % DO NOT CHANGE THIS
\usepackage[hyphens]{url}  % DO NOT CHANGE THIS
\usepackage{graphicx} % DO NOT CHANGE THIS
\usepackage{natbib}  % DO NOT CHANGE THIS AND DO NOT ADD ANY OPTIONS TO IT
\usepackage{caption} % DO NOT CHANGE THIS AND DO NOT ADD ANY OPTIONS TO IT
\usepackage{makecell}
\usepackage{algorithm}
\usepackage{algorithmic}
\usepackage{amsmath}
\usepackage{amssymb}
\usepackage{amsfonts}
\usepackage{multirow}
\usepackage{booktabs}
\usepackage{xcolor}
\usepackage{colortbl}
\definecolor{mred}{RGB}{238, 34, 12}
\definecolor{mgreen}{RGB}{1, 127, 0}
\definecolor{mblue}{RGB}{0, 77, 158}
\definecolor{darkgreen}{RGB}{0, 100, 0} % 通过RGB定义深绿色
\newcommand{\mredbf}[1]{\textcolor{mred}{\textbf{#1}}}
\newcommand{\mbluebf}[1]{\textcolor{mblue}{\textbf{#1}}}
\usepackage{arydshln} % 载入虚线表格的包
\usepackage{pifont}
\usepackage{newfloat}
\usepackage{listings}
\DeclareCaptionStyle{ruled}{labelfont=normalfont,labelsep=colon,strut=off} % DO NOT CHANGE THIS
\floatstyle{ruled}
\newfloat{listing}{tb}{lst}{}
\floatname{listing}{Listing}
\title{MIEScore: Human-Aligned Evaluation for Multi-Source Image Editing}
\author{
Zitong Xu\textsuperscript{1,2}, Huiyu Duan\textsuperscript{1}\thanks{Corresponding Authors.}, Xinyun Zhang\textsuperscript{3}, Weifei Xiong\textsuperscript{3}, Tianyi Zheng\textsuperscript{2}, Xiongkuo Min\textsuperscript{1}, \\Qiang Hu\textsuperscript{1}, Zhengxue Cheng\textsuperscript{1}, Bo Li\textsuperscript{2}\footnotemark[1], Guangtao Zhai\textsuperscript{1}\footnotemark[1] \\
}
\affiliations{
    \textsuperscript{1}Shanghai Jiao Tong University\\
    xuzitong@sjtu.edu.cn\\
    \textsuperscript{2}Vivo Mobile Communication Co., Ltd\\
    \textsuperscript{3}University of Electronic and Science Technology of China
}

\usepackage{bibentry}
\begin{document}
\setlength{\textfloatsep}{3.5pt plus 2pt minus 2pt}
\setlength{\dbltextfloatsep}{3.5pt plus 2pt minus 2pt}
\setlength{\dblfloatsep}{3.5pt plus 2pt minus 2pt}
\setlength{\intextsep}{3.5pt plus 2pt minus 2pt}
\setlength{\abovecaptionskip}{3.5pt plus 2pt minus 2pt}
\setlength{\belowcaptionskip}{3.5pt plus 2pt minus 2pt}
\setlength{\abovedisplayskip}{3pt}
\setlength{\belowdisplayskip}{3pt}
\setlength{\abovedisplayskip}{3pt}   % 上方间距
\setlength{\belowdisplayskip}{3pt}   % 下方间距

\maketitle
\begin{abstract}
Recent advances in unified multimodal models have significantly improved text-guided image editing abilities. In particular, models such as Nano-Banana-Pro and GPT-Image-2 demonstrate emerging capabilities in multi-source image editing (MIE), including tasks such as object synthesis, person-background composition, and cross-image style fusion. However, existing benchmarks and image editing assessment (IEQA) methods remain primarily focused on single-image editing tasks and largely overlook the more challenging setting of MIE. This highlights the urgent need for a comprehensive and human-aligned benchmark for MIE. To this end, we introduce \textbf{MIE-Bench}, the first large-scale multiple image editing benchmark with fine-grained human preference annotations. Specifically, MIE-Bench includes 3,000 editing instances across 16 tasks, each involving more than two source images and an editing prompt, together with 36K edited images produced by 12 state-of-the-art editing models and over 108K mean opinion scores (MOSs) covering visual quality, instruction following, and attribute preservation. Based on MIE-Bench, we propose \textbf{MIEScore}, a multimodal large language model (MLLM)-based evaluation model enhanced with skill optimization and multi-dimensional supervised fine-tuning, to provide human-aligned feedback for MIE. Extensive experiments show that MIEScore achieves state-of-the-art performance in aligning with human preferences and generalizes well across other IEQA datasets. Both the dataset and the model are available at \textcolor{mblue}{https://github.com/IntMeGroup/MIEScore}.
\end{abstract}

% Uncomment the following to link to your code, datasets, an extended version or similar.
% You must keep this block between (not within) the abstract and the main body of the paper.
% \begin{links}
%     \link{Code}{https://aaai.org/example/code}
%     \link{Datasets}{https://aaai.org/example/datasets}
%     \link{Extended version}{https://aaai.org/example/extended-version}
% \end{links}

\section{Introduction}
Earlier text-guided image editing (TIE) methods are primarily limited to single-source image settings \cite{ip2p,Magicbrush,ACE}. With the development of unified multimodal models, recent state-of-the-art image editing models such as Nano-Banana-Pro \cite{nanobanana}, Seedream5.0 \cite{bytedance_seedream5}, and Qwen-Image-Edit \cite{qwenedit} are now capable of handling multi-source image editing (MIE) tasks, such as object synthesis, person-background composition, and cross-image style fusion. However, the inherent complexity of multi-image editing introduces new challenges, including preserving subject identities, complex instruction following, and spatial reasoning across images \cite{wu2026miconbench}. However, most existing image editing benchmarks \cite{EditVal,ESurvey,I2EBench,IEBench,lmm4edit,editscore,editreward} are designed for single-source image editing. Although some works \cite{wang2025i2i,wu2026miconbench} have explored multi-source scenarios, they are still limited in task diversity, coverage of recent advanced models, and alignment with human preferences, as shown in Tabel~\ref{comparison}.

\begin{table}[t]
% \fontsize{7.5}{8}\selectfont
\centering
\caption{Comparison with existing multi-image editing benchmarks.
OmniContext~\cite{omnigen2},
XVerseBench~\cite{chen2025xverseconsistentmultisubjectcontrol},
I2I-Bench~\cite{wang2025i2i},
and MICON-Bench~\cite{wu2026miconbench}.}
 \resizebox{0.48\textwidth}{!}{
\begin{tabular}{lcccccccccc}
\toprule
\noalign{\vspace{-1.5pt}}
Databases& MOSs  & Instances &Edited Images & Models & Tasks \\
\hline
\noalign{\vspace{1.5pt}}
OmniContext  & \color{red}{\ding{55}} & 150  & 1,200 & 8 & 3 \\
XVerseBench  & \color{red}{\ding{55}} & 300  & 2,100 & 7 & 3 \\
I2I-Bench    & \color{red}{\ding{55}} & 500  & 2,000 & 4 & 5 \\
MICON-Bench  & \color{red}{\ding{55}} & 1,043 & 9,387 & 9 & 6 \\
\hline
\noalign{\vspace{1.5pt}}
  \rowcolor{gray!20}  % 这里添加灰色阴影
\textbf{MIE-Bench (Ours)}&\textbf{108K}&\textbf{3,000}&\textbf{36,000}&\textbf{12}&\textbf{16} \\
 \noalign{\vspace{-1.5pt}}
\bottomrule
\label{comparison}
\end{tabular}}
\end{table}
Beyond benchmarks, existing image editing quality assessment (IEQA) methods are not well-suited to multi-source conditions. Traditional image quality assessment (IQA) metrics \cite{MSSIM,SSIM,LPIPS} mainly focus on low-level visual quality but fail to capture semantic alignment with editing instructions. Vision-language approaches such as CLIPScore \cite{clipscore} and HPSv2 \cite{HPS} have shown notable progress in evaluating image generation by incorporating human visual feedback \cite{clipscore,blip,HPS,llavascore,imagereward}, but they mainly assess text-image alignment and are not designed to capture complex semantic transformations required in MIE. Moreover, both IQA metrics and vision-language approaches do not support multiple reference images, making them unsuitable for MIE evaluation. More recently, several studies have explored leveraging MLLMs for image editing evaluation \cite{lmm4edit,editscore,editreward}. Nevertheless, zero-shot MLLMs often exhibit weak alignment with human preferences \cite{lmm4edit}, while fine-tuned IEQA-specialized MLLMs \cite{editscore,editreward,xu2026edithf1mmillionscalerichhuman} are typically trained on single-source image editing datasets and thus cannot naturally handle multi-source inputs.

\begin{figure*}
    \centering
    \includegraphics[width=1\linewidth]{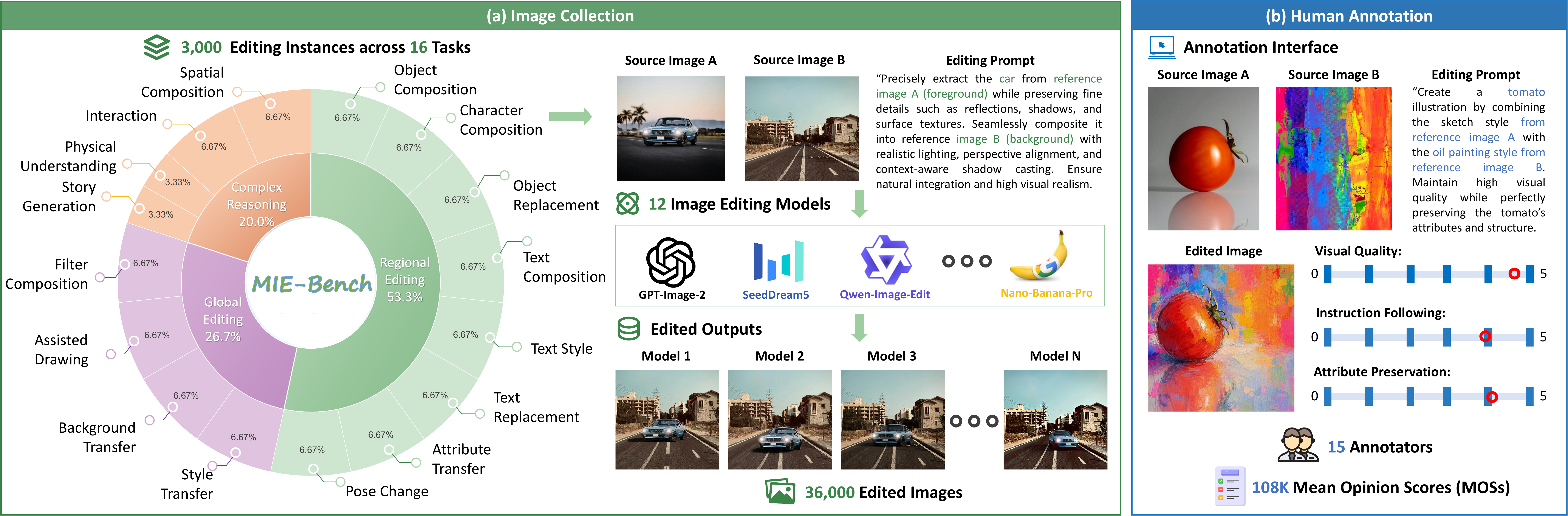}
    \caption{Overview of the construction pipeline of MIE-Bench. (a) Image collection. (b) Human annotation.}
    \label{benchmark}
\end{figure*}
In this work, we introduce \textbf{MIE-Bench}, a large-scale multi-source image editing evaluation dataset curated by trained annotators under a rigorous and standardized annotation protocol. MIE-Bench consists of \textbf{3,000} multi-source editing instances covering \textbf{16 editing tasks}, each with two or more source images collected from free photography website, as shown in Figure~\ref{benchmark}. We then generate \textbf{36K} edited images using \textbf{12} state-of-the-art image editing models, including both open-sourced models such as Qwen-Image-Edit \cite{qwenedit} and OmniGen2 \cite{omnigen2}, and closed-source models such as GPT-Image-2 \cite{openai_gptimage2} and Nano-Banana-Pro \cite{google_nanobananapro}. Through an extensive subjective study, we collect 1.62M human annotations evaluated from visual quality, instruction following and attribute preservation, which result in \textbf{108K} high-quality \textbf{mean opinion scores (MOSs)}. Building upon MIE-Bench, we propose MIEScore, the first human-aligned evaluation model for MIE, which jointly assesses visual quality, instruction following, and attribute preservation. It is built on an MLLM and trained via skill optimization and multi-dimensional supervised fine-tuning. Extensive experiments on MIE-Bench and other benchmarks demonstrate that MIEScore achieves state-of-the-art performance in multi-source image editing evaluation and also generalizes well to single-source image editing scenarios. 

The main contributions of this work include:
\begin{itemize}
\item We introduce MIE-Bench, a large-scale multi-source image editing benchmark with 36K edited images across diverse tasks, accompanied by comprehensive human annotations over multiple evaluation dimensions.
\item We propose MIEScore, the first human-aligned evaluation model for MIE, which jointly assesses visual quality, instruction following, and attribute preservation.
\item Extensive experiments on MIE-Bench and other benchmarks demonstrate that MIEScore achieves state-of-the-art performance in multi-source image editing evaluation and also generalizes well to single-source settings. 
\end{itemize}

\section{Related Work}
\subsection{Text-guided Image Editing}
Text-guided image editing has witnessed remarkable progress with the advent of large-scale diffusion models such as Stable Diffusion \cite{SD} and FLUX \cite{FLUX}. Early methods typically transform source image based on a target prompt \cite{Masactrl,cds,Flowedit}. Subsequent works adopt fine-tuning method \cite{ip2p,Magicbrush}, which allows models to modify style or semantic content directly from an instruction prompt. More recently, unified multimodal models \cite{omnigen2, nanobanana, seedream4} integrate understanding and generation within a single architecture, enabling flexible instruction-based editing and extending capabilities to multi-image settings such as style fusion and subject composition. However, challenges such as subject preservation and complex instruction understanding remain \cite{wu2026miconbench}.

\subsection{Benchmarks for Image Editing}
A growing number of benchmarks have been proposed for image editing \cite{han2025unireditbenchunifiedreasoningbasedimage,ye2025imgeditunifiedimageediting,wu2025krisbenchbenchmarkingnextlevelintelligent}, with increasing scale and richer task diversity. Beyond objective evaluation, some works focus on aligning assessment with human subjective perception \cite{I2EBench,wang2025i2i,xu2026edithf1mmillionscalerichhuman}, typically via pairwise comparisons or rating-based annotations. However, most existing benchmarks overlook multi-source image editing scenarios. Although recent efforts \cite{omnigen2,chen2025xverseconsistentmultisubjectcontrol,wu2026miconbench} begin to explore this setting, they remain limited in data scale and lack human preference annotations.

\subsection{Evaluation Metrics for Image Editing}
Traditional evaluation methods rely on image quality assessment (IQA) metrics, including full-reference (FR) IQA \cite{SSIM, LPIPS} for similarity measurement and no-reference (NR) IQA \cite{BRISQUE, NIQE} for visual quality assessment, but fail to capture editing instructions. Vision-language metrics \cite{imagereward,clipscore,HPS} have been proposed to evaluate image-text alignment, but they often struggle to assess semantic consistency between complex editing instructions and edited outputs. With the development of MLLMs, works such as \cite{lmm4edit, editscore, editreward,xu2026edithf1mmillionscalerichhuman} fine-tune MLLMs on human preference data to provide more accurate assessments. However, these evaluation methods typically only support single-source inputs and are not applicable to multi-source editing scenarios.

\section{MIE-Bench}
In this section, we introduce \textbf{MIE-Bench}, the first large-scale MIE dataset with fine-grained MOSs. The dataset comprises 3,000 high-quality MIE instances across 16 tasks, 36K edited images generated by 12 TIE models, and 108K MOSs covering visual quality, instruction following, and attribute preservation. With its diverse image content and human judgments, MIE-Bench provides a comprehensive benchmark for MIE evaluation aligned with human preferences.

\begin{figure*}[!t]
    \centering
    \includegraphics[width=1\linewidth]{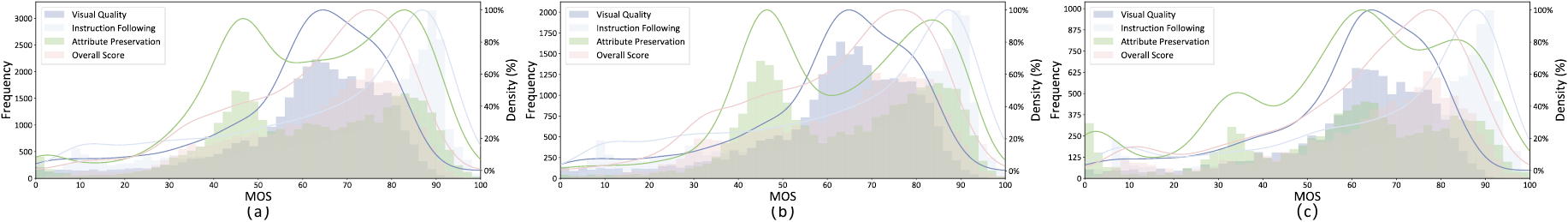}
    \caption{Distribution of MOSs in MIE-Bench. (a) All dataset. (b) Two source images. (c) Three or more source images.}
    \label{mosplot}
\end{figure*}
\begin{figure*}[!t]
    \centering
    \includegraphics[width=1\linewidth]{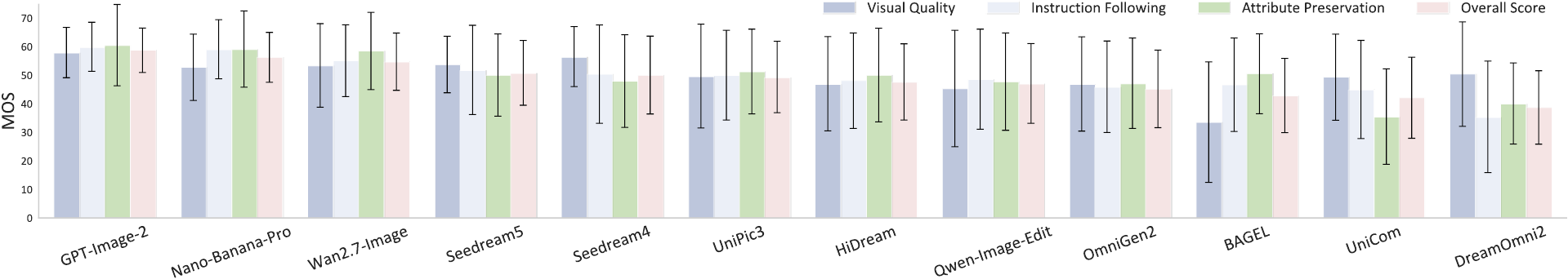}
    \caption{Comparison of image editing models on MIE-Bench across different dimensions.}
    \label{modelplot}
\end{figure*}
\begin{figure*}[!t]
    \centering
    \includegraphics[width=1\linewidth]{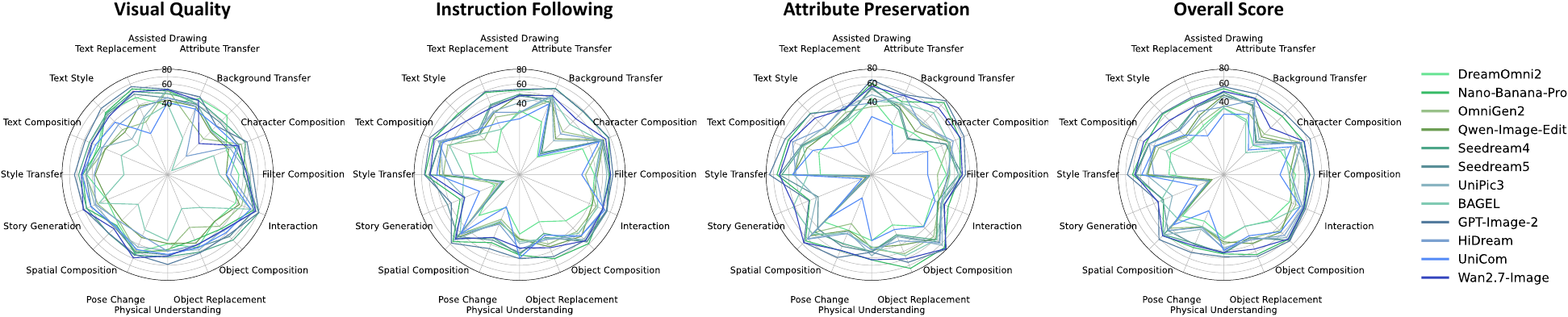}
    \caption{Comparison of image editing models across different editing tasks.}
    \label{catcomparison}
\end{figure*}

\subsection{Image Collection}
\begin{table}[t]
\setlength{\tabcolsep}{8pt} 
\centering
\caption{MIE models in our MIE-Bench.}
 \resizebox{0.48\textwidth}{!}{
\begin{tabular}{lcccccccccc}
\toprule
\noalign{\vspace{-1.5pt}}
Models& Released & Open Source \\
\hline
\noalign{\vspace{1.5pt}}
Seedream4 \cite{seedream4}&2025.09&\color{red}{\ding{55}}\\
Nano-Banana-Pro \cite{google_nanobananapro}&2025.11&\color{red}{\ding{55}}\\
Seedream5 \cite{seedream5}&2026.02&\color{red}{\ding{55}}\\
GPT-Image-2 \cite{openai_gptimage2}&2026.04&\color{red}{\ding{55}}\\
Wan2.7-Image \cite{mao2026wanimage}&2026.04&\color{red}{\ding{55}}\\
\midrule
BAGEL \cite{bagel}&2025.05&\textcolor{darkgreen}{\ding{51}}\\
OmniGen2 \cite{omni2}&2025.06&\textcolor{darkgreen}{\ding{51}}\\
DreamOmni2 \cite{dreamomni2}&2025.10&\textcolor{darkgreen}{\ding{51}}\\
Qwen-Image-Edit \cite{qwenedit}&2025.12&\textcolor{darkgreen}{\ding{51}}\\
UniPic3 \cite{wei2026skyworkunipic30unified}&2026.01&\textcolor{darkgreen}{\ding{51}}\\
UniCom \cite{zhao2026unicomunifiedmultimodalmodeling}&2026.03&\textcolor{darkgreen}{\ding{51}}\\
HiDream \cite{hidreamolimage}&2026.05&\textcolor{darkgreen}{\ding{51}}\\
 \noalign{\vspace{-1.5pt}}
\bottomrule
\label{MIE models}
\end{tabular}}
\end{table}

Considering both real-world usage and comprehensive evaluation of model capability, we select 16 MIE tasks, which are categorized into \textbf{global editing tasks}, \textbf{regional editing tasks}, and \textbf{complex reasoning tasks}, as shown in Figure~\ref{benchmark}. Specifically, global editing tasks involve modifications applied to the entire image, such as style transfer or filter composition; regional editing tasks focus on editing specific objects or regions, such as text replacement and attribute transfer; complex reasoning tasks involve high-level reasoning and world knowledge, such as spatial composition and story generation. Based on these editing tasks, we collect source images from publicly available photography websites and manually write simple editing instructions. We then leverage InternVL3.5~\cite{internvl3_5} to generate more explicit instructions, followed by manual verification and refinement. Finally, we generate a total of 36K edited images using 12 advanced MIE models as listed in Table~\ref{MIE models}.

\subsection{Subjective Experiment Setup}
To evaluate the edited images, we conduct a subjective quality assessment experiment using the MIE-Bench. This experiment is designed to capture human preferences for edited images, ensuring the results align with real-world human perception.

As shown in Figure~\ref{benchmark}(b), during the experiment, participants are presented with a set of source images, an edited image, and an editing instruction for each evaluation. Participants are asked to assess the edited image using a 5-point continuous scale from three aspects:
\begin{itemize}
    \item \textbf{Visual Quality (VQ):} focuses on assessing the overall quality of the edited images, considering factors such as the authenticity, distortion, color accuracy, and detail richness. 
    \item \textbf{Instruction Following (IF):} evaluates how accurately the edited image follows the given editing instructions, assessing whether the intended edits are correctly and completely applied.
   \item \textbf{Attribute Preservation (AP):} assesses the extent to which the edited image preserves key attributes and identity information from the source images.
\end{itemize}  

The experiment is conducted using a Python-based GUI displayed on a calibrated LED monitor with a resolution of 3840 $\times$ 2160, with images shown in 512$\times$512 resolution in a random order. Annotators are seated approximately two feet from the monitor in a controlled environment. Each image is evaluated by 15 participants. In total, we collect 1.62M rating scores across the three dimensions.

\subsection{Subjective Data Processing}
We follow the guidelines outlined in \cite{subject} to identify and exclude outliers, as well as to reject subjects who provide unreliable ratings. An individual rating for an image is considered an outlier if it falls outside $2$ standard deviations (if normal) or $\sqrt{20}$
standard deviations (if not normal) from the mean rating of that image. A subject is excluded if over $5\%$ of their ratings are outliers. 
As a result, no subject was excluded based on this criterion and $5.24\%$ of the total subjective ratings are removed. 
The remaining valid ratings are converted into Z-scores, then linearly scaled to the range [0,100]. The final MOS is calculated as follows
\begin{equation}
    z_{ij} = \frac{r_{ij} - \mu_i}{\sigma_i}, \ z_j = \frac{1}{N_j} \sum_{i=1}^{N_j} z_{ij}, \
    MOS_j = \frac{100(z_j + 3)}{6}
\end{equation}
where where $r_{ij}$ is the raw rating given by the i-th subject to the j-th image, $\mu_i$ is the mean rating and $\sigma_i$ is the standard deviation provided by the i-th subject and $N_j$ is the number of valid ratings for the j-th image. Finally, we collected over 108K MOSs for visual quality, instruction following, and attribute preservation.

\subsection{Subjective Data Analysis}
To comprehensively benchmark the overall performance of MIE models, following \cite{lmm4edit,xu2026edithf1mmillionscalerichhuman}, we compute an overall score, denoted as $S_{\mathrm{overall}}$, by aggregating the visual quality score $S_{\mathrm{VQ}}$, the instruction following score $S_{\mathrm{IF}}$, and the attribute preservation score $S_{\mathrm{AP}}$ using the following weighted geometric mean:
\begin{equation} S_\mathrm{Overall}=S_\mathrm{VQ}^{0.3} \times S_{IF}^{0.4}\times S_\mathrm{AP}^{0.3} \label{overall_score} 
\end{equation}
The editing alignment score is assigned a higher weight to emphasize the importance of satisfying editing instructions. 

Figure~\ref{mosplot} illustrates the MOS distributions of MIE-Bench from three perspectives: the entire dataset, samples with two source images, and samples with three or more source images. These distributions provide an overview of human preferences across different evaluation dimensions and reveal the quality characteristics under different multi-source editing scenarios. It can be observed that the MOS distributions of the attribute preservation dimension exhibit multiple peaks, which may result from the different numbers of source images whose attributes are successfully preserved. Figure~\ref{modelplot} shows the results of MIE models in MIE-Bench. GPT-Image-2 \cite{gptimage} achieves the best performance across all evaluation dimensions, while UniPic3 \cite{wei2026skyworkunipic30unified} demonstrates the strongest performance among open-source MIE models. 

We further benchmark each model’s performance across different editing tasks. As shown in Figure~\ref{catcomparison}, among complex reasoning tasks, models exhibit particularly poor performance on story generation, indicating that current MIE models still struggle with capturing the coherent relationships and visual transitions among multiple images. For global editing tasks, models encounter difficulties in background transfer, as they often fail to seamlessly integrate the foreground into the new background while maintaining visual coherence, suggesting that global scene-level understanding and content composition remain challenging. For regional-level tasks, text replacement achieves relatively lower performance because models have difficulty preserving the original typography and accurately reconstructing text details after editing, highlighting the limitations of current models in fine-grained attribute preservation.

\begin{figure*}[t]
    \centering
    \includegraphics[width=1\linewidth]{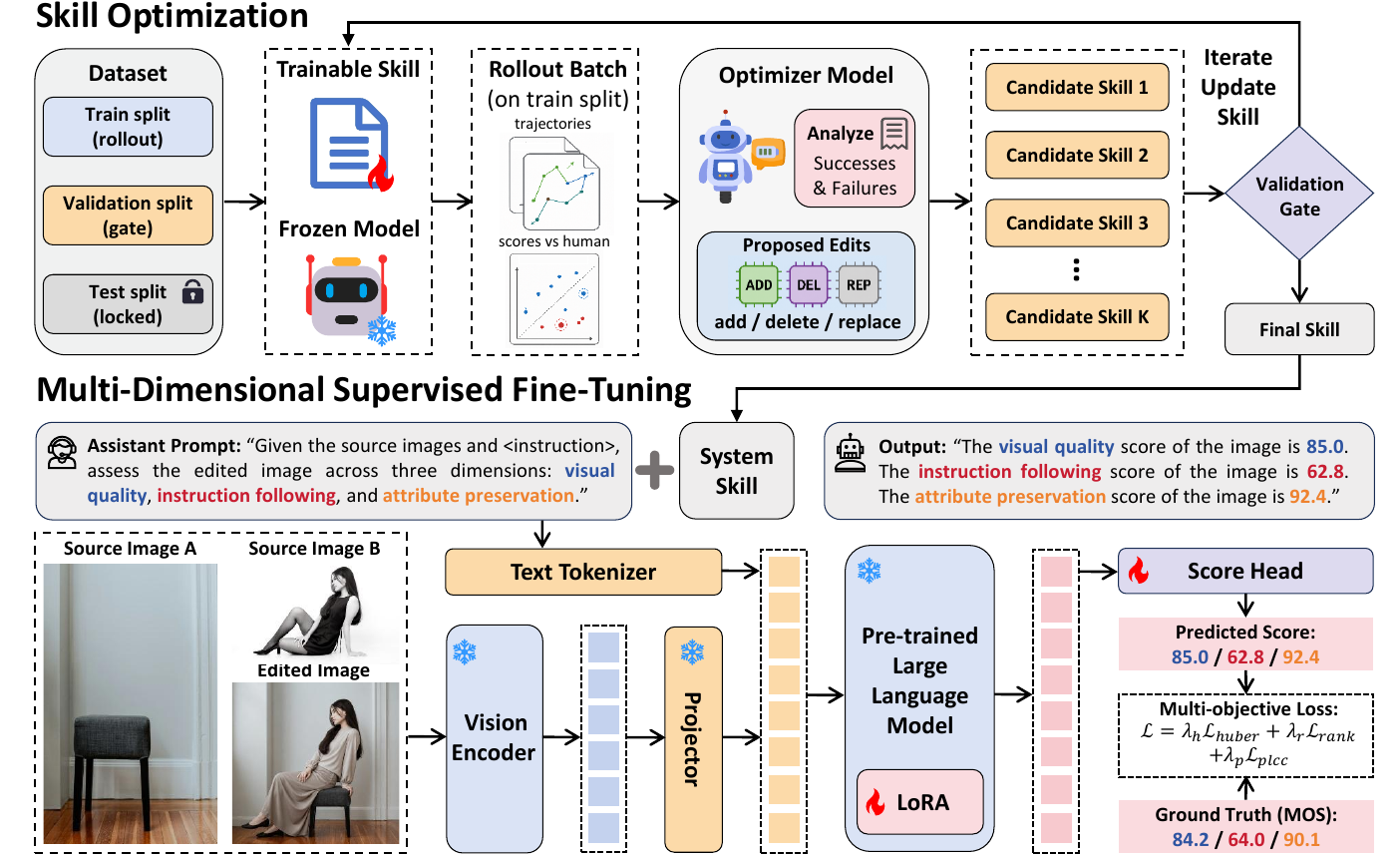}
    % \caption{Overview of the EditHF architecture and the editing model refinement process guided by EditHF. EditHF adopts an MLLM-based architecture that jointly processes the source image, edited image, and editing prompt. Visual features are extracted using a pretrained vision encoder, projected into the language embedding space, and fused with textual tokens through a multimodal backbone for joint reasoning. The resulting hidden representations are fed into an MLP head to predict fine-grained quality scores. The model is trained in three stages, namely textual learning, pointwise learning, and pairwise learning, which progressively establish a coarse to fine understanding of image editing quality. For model refinement, edited images generated by different editing models are evaluated and ranked by EditHF to construct preference pairs, which are then used to further optimize the editing models through direct preference optimization.}
    \caption{Overview of the training pipeline of MIEScore, consisting of two stages: Skill Optimization for evaluation criterion refinement and multi-dimensional supervised fine-tuning for human-aligned score prediction.}
    \label{model}
\end{figure*}

\section{MIEScore}
In this section, we present MIEScore, a unified evaluation model for multi-source image editing that jointly assesses visual quality, instruction following, and attribute preservation, aligning with human perceptual judgments.
\subsection{Model Architecture}
Due to the success of multimodal MLLMs in multimodal reasoning, we adopt an MLLM as the backbone of the evaluation model. Given an edited image $I_e$, a set of source images $\{I_s^i\}_{i=1}^{N}$, and an editing instruction $T$, the model predicts scores across three dimensions: visual quality, instruction following, and attribute preservation.

The edited image and source images are first encoded by a pre-trained vision encoder $E_v$, which converts them into visual tokens. These tokens are then combined with the tokenized instruction $T_p$ to form a unified multimodal sequence. The resulting sequence is fed into a large language model $H_\psi$ for joint cross-modal and cross-source reasoning, where LoRA module is applied to enable efficient adaptation while preserving pretrained knowledge. The final hidden state is denoted as:
\begin{equation}
\mathbf{h} = H_\psi\big(E_v(I_e), E_v(I_s^{1:N}), T_p\big).
\end{equation}

To produce fine-grained evaluation scores, a lightweight regression head $S_\omega$ processes the hidden state to predict a three-dimensional score vector.
\begin{equation}
\hat{\mathbf{y}} =
S_\omega(\mathbf{h}) =
\big[\hat{y}^{VQ}, \hat{y}^{IF}, \hat{y}^{AP}\big],
\end{equation}
corresponding to visual quality, instruction following, and attribute preservation.
\begin{table*}[!t]
\setlength{\tabcolsep}{8pt} 
\caption{Performance comparisons of quality evaluation methods on MIE-Bench. 
SRCC ($\rho_s$), KRCC ($\rho_k$), and PLCC ($\rho_p$) metrics are reported. 
$\heartsuit$ NR IQA metrics, $\spadesuit$ FR IQA metrics, $\diamondsuit$ vision-language methods, $\clubsuit$ general-purpose MLLMs,  \ding{73} IEQA-specialized MLLMs. The fine-tuned results are marked with \raisebox{0.5ex}{\scriptsize \ding{91}}. The best results are highlighted in \mredbf{red}, and the second-best results are highlighted in \mbluebf{blue}.}
\begin{center}
\centering
\renewcommand\arraystretch{0.95}
\centering
\belowrulesep=0pt
\aboverulesep=0pt
\setlength{\tabcolsep}{12pt} 
% \fontsize{3}{3.2}\selectfont
\resizebox{1\textwidth}{!}{
\begin{tabular}{l||ccc|ccc|ccc||ccc}
\toprule
\noalign{\vspace{1.5pt}}
Dimensions& \multicolumn{3}{c}{Visual Quality} & \multicolumn{3}{c}{Instruction Following} & \multicolumn{3}{c}{Attribute Preservation}& \multicolumn{3}{c}{Overall Score} \\
\cmidrule(lr){2-4}
\cmidrule(lr){5-7}
\cmidrule(lr){8-10}
\cmidrule(lr){11-13}
\noalign{\vspace{-1.5pt}}
Methods/Metrics & $\rho_s$ & $\rho_k$ & $\rho_p$ &  $\rho_s$ & $\rho_k$ & $\rho_p$ & $\rho_s$ & $\rho_k$ & $\rho_p$& $\rho_s$ & $\rho_k$ & $\rho_p$  \\
% Method/Metrics & SRCC & KRCC & PLCC &  SRCC & KRCC & PLCC  & SRCC & KRCC & PLCC   \\
\midrule
\noalign{\vspace{1pt}}
$\heartsuit$DIIVINE \cite{DIIVINE} &0.1284&0.0817&0.3412&0.0526&0.0289&0.1103&0.0185&0.0102&0.1934&0.0856&0.0415&0.1806\\
$\heartsuit$BRISQUE \cite{BRISQUE} &0.3186&0.2194&0.3728&0.1421&0.0897&0.1493&0.1216&0.0825&0.2018&0.1216&0.0825&0.2018\\
$\heartsuit$CNNIQA \cite{CNN}\raisebox{0.5ex}{\scriptsize \ding{91}} &0.5127&0.4318&0.5415&0.2294&0.1612&0.2410&0.2958&0.2714&0.2186&0.2958&0.2714&0.2186\\
$\heartsuit$HyperIQA \cite{Hyper}\raisebox{0.5ex}{\scriptsize \ding{91}} &0.6712&0.4826&0.6894&0.2417&0.1689&0.2553&0.3024&0.2016&0.3187&0.3024&0.2016&0.3187\\
$\heartsuit$TOPIQ \cite{TOPIQ}\raisebox{0.5ex}{\scriptsize \ding{91}} &0.6815&0.5893&0.6968&0.3782&0.2641&0.4016&0.6487&0.6812&0.4729&0.4487&0.3812&0.4729\\
\midrule
\noalign{\vspace{1pt}}
$\spadesuit$SSIM \cite{SSIM} &0.1568&0.0821&0.1187&0.1632&0.1095&0.2217&0.1512&0.1094&0.1726&0.1512&0.1094&0.1726\\
$\spadesuit$MSSIM \cite{MSSIM} &0.0821&0.0317&0.0889&0.1225&0.1214&0.1368&0.1032&0.0815&0.1714&0.1032&0.0815&0.1714\\
$\spadesuit$SCSSIM \cite{SCSSIM} &0.1463&0.1258&0.1921&0.1127&0.0416&0.1383&0.2129&0.1273&0.2286&0.2129&0.1273&0.2286\\
$\spadesuit$LPIPS \cite{LPIPS} &0.2015&0.1382&0.2864&0.1318&0.0556&0.1079&0.2413&0.1734&0.2621&0.2413&0.1734&0.2621\\
$\spadesuit$Q-Align \cite{qalign}\raisebox{0.5ex}{\scriptsize \ding{91}} &\mbluebf{0.7834}&0.6152&0.7928&0.3517&0.2729&0.3836&0.7885&0.6361&0.7718&0.6085&0.5361&0.6218\\
\midrule
\noalign{\vspace{1pt}}

$\diamondsuit$CLIPScore \cite{clipscore} &0.2337&0.1584&0.2419&0.2216&0.1498&0.2325&0.2451&0.1663&0.2714&0.2451&0.1663&0.2714\\
$\diamondsuit$ImageReward \cite{imagereward} &0.4218&0.2896&0.4527&0.3014&0.2053&0.3312&0.2186&0.1861&0.2795&0.2186&0.1861&0.2795\\
$\diamondsuit$HPSv2 \cite{HPS} &0.4512&0.3215&0.4736&0.3294&0.2286&0.3621&0.2618&0.1627&0.2889&0.2618&0.1627&0.2889\\
$\diamondsuit$VQAScore \cite{vqa} &0.3189&0.2147&0.3295&0.2916&0.1964&0.2783&0.2317&0.1528&0.2654&0.2317&0.1528&0.2654\\
\midrule
\noalign{\vspace{1pt}}

$\clubsuit$mPLUG-Owl3 (7B) \cite{mplug} &0.3324&0.2418&0.2976&0.0721&0.0513&0.0348&0.1689&0.1186&0.0863&0.1832&0.1053&0.0742\\
$\clubsuit$MiniCPM-V2.6 (8B) \cite{minicpm} & 0.0499 & 0.0366 & 0.0215 & 0.1120 & 0.0803 & 0.1533 & 0.0870 & 0.0615 & 0.1447&0.1377	&0.0954	&0.1611 \\
$\clubsuit$Ovis2.5 (8B) \cite{ovis25} & 0.1558 & 0.1199 & 0.0683 & 0.2081 & 0.1587 & 0.2472 & 0.0504 & 0.0390 & 0.1104&0.0774	&0.0541	&0.2182 \\
$\clubsuit$Qwen3-VL (7B) \cite{qwen3} & 0.1844 & 0.1384 & 0.1379 & 0.2642 & 0.1986 & 0.3934 & 0.0865 & 0.0616 & 0.2781&0.2114	&0.1456	&0.3678 \\
$\clubsuit$InternVL3.5 (8B) \cite{internvl3_5} & 0.1898 & 0.1424 & 0.1804 & 0.1309 & 0.1006 & 0.1898 & 0.0082 & 0.0050 & 0.0712&0.0810	&0.0554	&0.1782 \\
$\clubsuit$DeepSeekVL2 (small) \cite{deepseekv2} & 0.0681 & 0.0542 & 0.1018 & 0.1146 & 0.0893 & 0.0687 & 0.0854 & 0.0632 & 0.0247 &0.0760	&0.0568	&0.0114\\
$\clubsuit$Gemini-3.1-Pro \cite{gemini3pro2026} & 0.5372 & 0.4223 & 0.4473 & 0.4509 & 0.3461 & 0.3358 & 0.4261 & 0.3217 & 0.3671 &0.5244	&0.3728	&0.3314\\
$\clubsuit$ChatGPT-5 \cite{gpt5} & 0.5416 & 0.4275 & 0.3512 & 0.5032 & 0.4072 & 0.3953 & 0.4335 & 0.3511 & 0.3712 &0.5349	&0.4072	&0.3892\\
$\clubsuit$DeepSeekVL2 (small)\raisebox{0.5ex}{\scriptsize \ding{91}} \cite{deepseekv2} & 0.7825 &  \mbluebf{0.6229} & 0.8062 & \mbluebf{0.8610} & \mbluebf{0.6588} & \mbluebf{0.8894} & 0.8623 & 0.6754 & 0.8551 &\mbluebf{0.8872}	&\mbluebf{0.7116}	&0.8444\\
$\clubsuit$InternVL3.5 (8B)\raisebox{0.5ex}{\scriptsize \ding{91}} \cite{internvl2}& 0.7609 & 0.5693 & \mbluebf{0.8099}
& 0.8288 & 0.6332 & 0.8657 
& \mbluebf{0.8727} & \mbluebf{0.6805} & \mbluebf{0.8782} &0.8841	&0.6977	&\mbluebf{0.8654}\\
\midrule
\noalign{\vspace{1pt}}

\ding{73}LMM4Edit \cite{lmm4edit} & 0.4888 & 0.3428 & 0.5928 & 0.6104 & 0.4333 & 0.6326 & 0.3798 & 0.2638 & 0.4377&0.5822	&0.4131	&0.6209 \\
\ding{73}EditScore (Qwen3) \cite{editscore} & 0.2461 & 0.1678 & 0.3050 & 0.3585 & 0.2454 & 0.3367 & 0.2736 & 0.1858 & 0.2939 &0.2816	&0.1888	&0.3159\\
\ding{73}EditReward (MiMo) \cite{editreward} & 0.3423 & 0.2339 & 0.3743 & 0.4942 & 0.3421 & 0.4944 & 0.3454 & 0.2350 & 0.3461 &0.4641	&0.3164	&0.4387\\
\ding{73}EditHF \cite{xu2026edithf1mmillionscalerichhuman} & 0.4941 & 0.3478 & 0.6004 & 0.5426 & 0.3823 & 0.5414 & 0.4253 & 0.2942 & 0.4552 &0.4720	&0.3251	&0.5173\\

\midrule
\noalign{\vspace{1pt}}

\rowcolor{gray!20}
\textbf{MIEScore (Ours)} & \mredbf{0.8239} & \mredbf{0.6395} & \mredbf{0.8257} & \mredbf{0.8703} & \mredbf{0.6985} & \mredbf{0.9077} & \mredbf{0.9006} & \mredbf{0.7306} & \mredbf{0.9075} &\mredbf{0.9000}	&\mredbf{0.7283}	&\mredbf{0.8954}\\
\bottomrule

\end{tabular}}

\label{performances}
\end{center}
\end{table*}

\begin{table*}[t]
\belowrulesep=0pt
\aboverulesep=0pt
\centering
\renewcommand{\arraystretch}{0.85}
\caption{Comparisons of the alignment between different evaluation methods and human perception in evaluating image editing models. The best results are highlighted in \mredbf{red}, and the second-best results are highlighted in \mbluebf{blue}. \raisebox{0.5ex}{\scriptsize \ding{91}} denotes fine-tuned models.} 
\setlength{\tabcolsep}{4pt} 
\resizebox{\textwidth}{!}{
\begin{tabular}{l||c:ccc| c:ccc| c:ccc| c:cc| c:c}
\toprule
\noalign{\vspace{1.5pt}}
Dimensions & \multicolumn{4}{c}{Visual Quality} & \multicolumn{4}{c}{Instruction Following} & \multicolumn{4}{c}{Attribute Preservation}& \multicolumn{3}{c}{Overall Score}& \multicolumn{2}{c}{Overall Rank}\\
\cmidrule(lr){2-5}
\cmidrule(lr){6-9}
\cmidrule(lr){10-13}
\cmidrule(lr){14-16}
\cmidrule(lr){17-18}

\noalign{\vspace{1.5pt}}
Models/Metrics & Human &\cellcolor{gray!20}Ours\raisebox{0.5ex}{\scriptsize \ding{91}} & DeepSeekVL2\raisebox{0.5ex}{\scriptsize \ding{91}} & Gemini-3.1 & Human &\cellcolor{gray!20}Ours\raisebox{0.5ex}{\scriptsize \ding{91}}  &DeepSeekVL2\raisebox{0.5ex}{\scriptsize \ding{91}}&LMM4Edit &Human&\cellcolor{gray!20}Ours\raisebox{0.5ex}{\scriptsize \ding{91}}&InternVL3.5\raisebox{0.5ex}{\scriptsize \ding{91}}&ChatGPT-5&Human&\cellcolor{gray!20}Ours\raisebox{0.5ex}{\scriptsize \ding{91}} &Gemini-3.1&Human&\cellcolor{gray!20}Ours\\

\hline
\noalign{\vspace{1.5pt}}
GPT-Image-2 & 57.88 & \cellcolor{gray!20}56.89 & 65.67 & 74.05 & 59.93 &\cellcolor{gray!20} 57.49 & 65.80 & 59.68 & 60.51 &\cellcolor{gray!20} 55.04 & 65.35 & 85.63 & 58.70 & \cellcolor{gray!20}55.97  & 71.36 &1&\cellcolor{gray!20}1\\
Nano-Banana-Pro & 52.72 &\cellcolor{gray!20} 52.91 & 61.35 & 71.40 & 59.04 &\cellcolor{gray!20} 55.63 & 64.51 & 58.12 & 59.10 &\cellcolor{gray!20} 54.56 & 63.53 & 84.98 & 56.18 &\cellcolor{gray!20} 53.78 & 70.30 &2&\cellcolor{gray!20}2\\
Wan2.7-Image & 53.38 &\cellcolor{gray!20} 53.47 & 61.91 & 71.65 & 55.05 & \cellcolor{gray!20}53.08 & 61.60 & 57.63 & 58.45 &\cellcolor{gray!20} 52.90 & 62.93 & 84.63 & 54.50 &\cellcolor{gray!20} 52.06 & 68.99 &3&\cellcolor{gray!20}3\\
Seedream5 & 53.71 &\cellcolor{gray!20} 55.65 & 63.28 & 71.20 & 51.82 &\cellcolor{gray!20} 48.59 & 61.70 & 58.16 & 50.00 & \cellcolor{gray!20}45.75 & 56.99 & 84.25 & 50.74 & \cellcolor{gray!20}48.60 & 67.37 &4&\cellcolor{gray!20}4\\
Seedream4 & 56.48 &\cellcolor{gray!20} 55.43 & 63.97 & 72.30 & 50.32 & \cellcolor{gray!20}47.01 & 57.69 & 57.28 & 47.91 & \cellcolor{gray!20}44.47 & 54.78 & 82.52 & 49.99 &\cellcolor{gray!20}  47.07 & 66.97 &5&\cellcolor{gray!20}6\\
UniPic3 & 49.64 &\cellcolor{gray!20} 50.97 & 58.71 & 70.92 & 49.97 & \cellcolor{gray!20}47.67 & 56.66 & 57.01 & 51.25 & \cellcolor{gray!20}47.60 & 57.10 & 83.90 & 48.70 &\cellcolor{gray!20} 47.40 & 67.64 &6&\cellcolor{gray!20}5\\
HiDream & 46.98 &\cellcolor{gray!20} 49.28 & 55.55 & 68.57 & 48.04 &\cellcolor{gray!20} 45.11 & 54.40 & 55.72 & 50.01 &\cellcolor{gray!20} 46.43 & 54.82 & 82.00 & 46.91 &\cellcolor{gray!20} 45.24  & 65.90 &7&\cellcolor{gray!20}7\\
Qwen-Image-Edit & 45.32 &\cellcolor{gray!20} 50.26 & 54.66 & 68.07 & 48.57 & \cellcolor{gray!20}44.65 & 52.50 & 55.03 & 47.72 & \cellcolor{gray!20}45.01 & 52.76 & 81.28 & 46.21 &\cellcolor{gray!20} 44.84 & 65.83 &8&\cellcolor{gray!20}8\\
OmniGen2 & 46.88 &\cellcolor{gray!20} 48.46 & 54.46 & 69.03 & 45.93 & \cellcolor{gray!20}43.49 & 53.09 & 55.40 & 47.16 & \cellcolor{gray!20}44.00 & 51.45 & 82.62 & 44.85 & \cellcolor{gray!20}43.57  & 65.25 &9&\cellcolor{gray!20}9\\
UniCom & 49.28 &\cellcolor{gray!20} 52.51 & 54.36 & 70.68 & 44.97 & \cellcolor{gray!20}39.43 & 49.97 & 55.55 & 35.46 &\cellcolor{gray!20} 33.51 & 41.37 & 80.13 & 41.67 &\cellcolor{gray!20} 39.34 & 64.76 &10&\cellcolor{gray!20}11\\
BAGEL & 33.51 &\cellcolor{gray!20} 39.70 & 44.78 & 59.67 & 46.63 &\cellcolor{gray!20} 43.09 & 51.30 & 52.89 & 50.47 & \cellcolor{gray!20}45.65 & 54.33 & 82.03 & 41.19 & \cellcolor{gray!20}41.18 & 59.72 &11&\cellcolor{gray!20}10\\
DreamOmni2 & 50.35 &\cellcolor{gray!20} 55.03 & 55.20 & 68.02 & 35.38 &\cellcolor{gray!20} 31.31 & 42.12 & 49.39 & 40.06 &\cellcolor{gray!20} 36.82 & 45.28 & 76.68 & 38.26 &\cellcolor{gray!20} 36.27 & 60.06 &12&\cellcolor{gray!20}12\\

\hline
\noalign{\vspace{1.5pt}}
SRCC to human $\uparrow$&&\cellcolor{gray!20}\mredbf{0.944}&
\mbluebf{0.923}&0.867&&\cellcolor{gray!20}\mredbf{0.979}&\mbluebf{0.958}&0.895&&\cellcolor{gray!20}\mredbf{0.972}&\mbluebf{0.951}&0.832&&\cellcolor{gray!20}\mredbf{0.986}& \mbluebf{0.972}&&\cellcolor{gray!20}\mredbf{0.986}\\
\bottomrule
\end{tabular}
}
\label{compare_TIE}
\end{table*}
\subsection{Skill Optimization}
To help the model better understand the evaluation criteria and develop a more consistent assessment process, we first perform skill optimization inspired by SkillOpt~\cite{yang2026skillopt}. We optimize a textual evaluation skill $S$, which defines the evaluation rubric, grounding rules, and scoring principles.

Given a multi-image evaluation sample $x = \{I_s^{1:N}, I_e, T\}$, the evaluator produces a three-dimensional score vector:
\begin{equation}
\hat{\mathbf{y}} = f(S, x),
\end{equation}
where $f(\cdot)$ is a frozen multimodal evaluator conditioned on the skill $S$.

Starting from an initial skill $S^{(0)}$, we perform iterative refinement. 
At each iteration $t$, the frozen evaluator executes rollouts with the current skill, 
producing a set of evaluation trajectories $\mathcal{T}^{(t)}$ containing model predictions, 
human scores, and corresponding error signals. We then identify failure trajectories 
with large deviations from human annotations:
\begin{equation}
\mathcal{E}^{(t)} =
\{\tau \in \mathcal{T}^{(t)} \mid 
\| f(S^{(t)}, x_\tau)-\mathbf{y}_\tau \| \ \text{is large} \}.
\end{equation}
An optimizer model performs reflection over the current skill and identified failure cases. 
Based on the analysis of successful and failed evaluations, it generates bounded skill edits through 
add, delete, and replace operations, producing $K$ candidate skills:
\begin{equation}
S_k^{(t+1)} = \mathcal{O}(S^{(t)}, \mathcal{E}^{(t)}; e_k), \quad k = 1, \dots, K,
\end{equation}
where $\mathcal{O}(\cdot)$ denotes the optimizer model and $e_k$ represents the proposed edit operation.

Each candidate skill is evaluated on the validation set to estimate its expected ranking performance, and the best one is selected:
\begin{equation}
S^{(t+1)} = \arg\max_{S_k} \ \mathbb{E}_{(x, \mathbf{y}) \sim \mathcal{D}_{val}}
\left[
\mathrm{SRCC}\big(f(S_k, x), \mathbf{y}\big)
\right].
\end{equation}

We iteratively perform this process to derive an optimized skill that enables the model to better comprehend the evaluation criteria and produce more reliable assessments.

\subsection{Multi-Dimensional Supervised Fine-Tuning}

To enable the model to produce fine-grained evaluation scores aligned with human preferences, we further fine-tune the model using MOS annotations. We jointly optimize all three evaluation dimensions within a multi-objective learning framework. The overall training objective is defined as:
\begin{equation}
\mathcal{L} =
\lambda_h \mathcal{L}_{huber}
+
\lambda_r \mathcal{L}_{rank}
+
\lambda_p \mathcal{L}_{plcc}.
\end{equation}

Each loss is computed independently over the three evaluation dimensions $d \in \{VQ, IF, AP\}$. The Huber loss is applied for point-wise score regression, the ranking loss enforces pairwise consistency, and the PLCC loss promotes correlation alignment with human judgments.
\begin{align}
\mathcal{L}_{huber} &= \sum_{d} \text{Huber}(\hat{y}^d, y^d), \\
\mathcal{L}_{rank} &= \sum_{d} \mathcal{L}_{rank}^d, \\
\mathcal{L}_{plcc} &= \sum_{d} \big(1 - \rho(\hat{y}^d, y^d)\big),
\end{align}
where $\rho(\cdot)$ denotes Pearson correlation.

This formulation leads to a unified and human-aligned evaluation model.

\section{Experiments}
In this section, we evaluate the performance of our MIEScore through extensive experiments.
\subsection{Experiment Setup}
We select Qwen3-VL (8B) \cite{qwen3} as the backbone of MIEScore and train it on our MIE-Bench, which is split into training, validation, and testing sets with a ratio of 3:1:1. All comparison models are trained using the same data split for fair comparison. For methods that support only single-source image inputs, we compute the results for each source image separately and report the average score. All models are implemented with PyTorch and trained on a single NVIDIA RTX A6000 GPU.
\begin{table*}[tb]
\centering
\caption{Ablation study on the different backbones and training strategy.}
\label{ablation}

 \resizebox{1\textwidth}{!}{
\begin{tabular}{lcccc|cc|cc|cc|cc}
\toprule
 \noalign{\vspace{-1.3pt}}
\multicolumn{5}{c}{Backbone$\&$Strategy} & \multicolumn{2}{c}{Visual Quality} & \multicolumn{2}{c}{Instruction Following} & \multicolumn{2}{c}{Attribute Preservation}& \multicolumn{2}{c}{Overall Score}\\
 \noalign{\vspace{-1.0pt}}
\cmidrule(lr){1-5} \cmidrule(lr){6-7} \cmidrule(lr){8-9} \cmidrule(lr){10-11}  \cmidrule(lr){12-13}
 \noalign{\vspace{-1.5pt}}
Backbone&Score Head &LoRA & Skill Optimization & Supervised Fine-tuning& SRCC & PLCC & SRCC & PLCC & SRCC & PLCC & SRCC & PLCC\\ 
\hline
\noalign{\vspace{1pt}}

Qwen3-VL \cite{qwen3}& & \checkmark &  &\checkmark& 0.7428 & 0.7714 & 0.8013 & 0.8346 & 0.8195 & 0.8421 & 0.8132 & 0.8367\\

Qwen3-VL \cite{qwen3}&\checkmark &  &  & \checkmark  & 0.7165 & 0.7449 & 0.7820 & 0.8128 & 0.8017 & 0.8275 & 0.7954 & 0.8193\\

Qwen3-VL \cite{qwen3}&\checkmark & \checkmark  &  & \checkmark  & 0.8051 & 0.8136 & 0.8539 & 0.8762 & 0.8684 & 0.8719 & 0.8627 & 0.8758\\

Qwen3-VL \cite{qwen3} & & & \checkmark &   & 0.6889 & 0.7210 & 0.7542 & 0.7896 & 0.7713 & 0.8038 & 0.7645 & 0.7761\\

\rowcolor{gray!20}
Qwen3-VL \cite{qwen3}&\checkmark & \checkmark &  \checkmark &  \checkmark & \textbf{0.8239} & \textbf{0.8257} & \textbf{0.8703} & \textbf{0.9077} & \textbf{0.9006} & \textbf{0.9075} & \textbf{0.9000} & \textbf{0.8954}\\

Qwen2.5-VL \cite{qwenvl2}&\checkmark & \checkmark &  \checkmark &  \checkmark & 0.7821 & 0.8048 & 0.8426 & 0.8783 & 0.8789 & 0.8864 & 0.8650 & 0.8691\\

DeepSeekVL2 \cite{deepseekv2} &\checkmark & \checkmark &  \checkmark &  \checkmark &
0.7825 & 0.8062 & 0.8510 & 0.8894 & 0.8623 & 0.8551 & 0.8772 & 0.8844\\

InternVL3.5 \cite{internvl3_5}&\checkmark & \checkmark &  \checkmark&  \checkmark  & 
0.7609 & 0.8099 & 0.8288 & 0.8657 & 0.8727 & 0.8782 & 0.8741 & 0.8754 \\

 \noalign{\vspace{-1.5pt}}
\bottomrule
\end{tabular}}
\end{table*}
\begin{table*}[t]
\renewcommand\arraystretch{0.95}
\centering
\belowrulesep=0pt
\aboverulesep=0pt
\caption{Zero-shot evaluation results on other image editing benchmarks. The best results are highlighted in \mredbf{red}, and the second-best results are highlighted in \mbluebf{blue}. “-” indicates that the model has been trained on the corresponding dataset, and its performance is therefore omitted to ensure a fair cross-dataset comparison.}
\label{other_dataset}
\setlength{\tabcolsep}{9pt} 
 \resizebox{1\textwidth}{!}{
\begin{tabular}{l||ccccccc}
\toprule
\noalign{\vspace{1.5pt}}
Methods/Benchmarks&GenAI-Bench&AROURA-Bench &EditScore-Bench&ImagenHub &EBench-18K&EditReward-Bench& IEQA\\
Metrics& (Acc., \%)&(Acc., \%) &(Acc., \%) &(SRCC)&(SRCC)& (SRCC)&(SRCC)\\
\midrule
\noalign{\vspace{1pt}}
% GPT-4o \cite{chatgpt4o}& 53.54 & 50.81 & 67.30 & 0.3821 &0.4108&0.2831&0.3628\\
ChatGPT-5 \cite{gpt5}& 59.61 & 47.27 & 75.24 & 0.4085 &0.4260&0.3781&0.4175\\
% Gemini-2.0-Flash \cite{gemini2}& 53.32 & 44.31 & 69.58 & 0.2369 &0.3892&0.3347&0.3512\\
Gemini-2.5-Flash \cite{nanobanana}& 57.01 & 47.63 & 70.26 &  0.4162 &0.4337&0.3802&0.4258\\
Qwen3-VL (7B) \cite{qwen3}& 44.58 & 34.28 & 42.51 & 0.2426 &0.3726&0.2731&0.3860\\
InternVL3.5 (7B) \cite{internvl3_5}& 45.17 & 36.92 & 45.18 & 0.2071 &0.3511&0.2406&0.3091\\
EditScore (Qwen3) \cite{editscore}& 50.24 & 42.18 & - & 0.3062 &0.3206&0.2590&0.3805\\
EditReward (MiMo) \cite{editreward}&  65.72 & 63.62 & 71.25 & 0.3520 &0.3718&-&0.4643 \\
LMM4Edit \cite{lmm4edit} & 63.27 &  65.21 & 70.64 & 0.3726 &-&0.3958& \mbluebf{0.5012}\\
EditHF \cite{xu2026edithf1mmillionscalerichhuman}&\mbluebf{67.92}&\mbluebf{66.08}&\mredbf{77.50}&\mbluebf{0.4627}&\mbluebf{0.4526}&\mredbf{0.4325}&-\\
\midrule
\noalign{\vspace{1pt}}
 \rowcolor{gray!20}
\textbf{MIEScore (Ours)}&\mredbf{76.22}&\mredbf{70.28}& \mbluebf{76.15} & \mredbf{0.5508} &\mredbf{0.5024}& \mbluebf{0.4153}& \mredbf{0.5047}\\
\bottomrule
\end{tabular}}
\end{table*}
\subsection{Evaluation on MIE-Bench}
Table~\ref{performances} presents the performance of our MIEScore in comparison with NR IQA and FR IQA methods, vision-language methods, general-purpose MLLMs and IEQA-specialized MLLMs. Spearman Rank Correlation Coefficient (SRCC), Pearson Linear Correlation Coefficient (PLCC), and Kendall’s Rank Correlation Coefficient (KRCC) are reported.

NR IQA and FR IQA methods without fine-tuning perform poorly, as their features are primarily designed for conventional image distortions and are not well-suited to high-level artifacts. Fine-tuned FR IQA methods improve performance in visual quality and attribute preservation, but still struggle to reliably capture editing alignment. Vision-language methods also show limited effectiveness, as they are mainly optimized for text–image alignment and have limited capability in handling complex editing instructions. IEQA-specialized MLLMs can achieve reasonable performance in zero-shot settings, but still perform poorly overall due to training on single-source datasets, difficulty in multi-source scenarios, and the lack of native support for multi-source image inputs. General-purpose MLLMs exhibit suboptimal performance, particularly for open-source models, with a substantial gap compared to proprietary models. This is largely because MIE evaluation is a challenging task that requires strong multimodal understanding and reasoning abilities. In contrast, our MIEScore achieves the best alignment with human perception across all evaluation dimensions, while fine-tuned MLLMs also demonstrate strong performance, further validating the effectiveness and value of MIE-Bench.

Table~\ref{compare_TIE} further demonstrates the effectiveness of our model in benchmarking MIE models. Our model achieves the highest SRCC with human scores, indicating its strong ability to accurately reflect human preferences and evaluate MIE models. Moreover, the results reveal a clear performance gap between closed-source image editing methods and open-source models. Even the lowest-performing closed-source model, Seedream4, outperforms the best-performing open-source model, UniPic3, in terms of the overall score.

\subsection{Ablation Study}
To validate the effectiveness of different components and training strategies in MIEScore, we conduct comprehensive ablation studies, with the results summarized in Table~\ref{ablation}. The comparisons among the first three rows highlight the necessity of the score head and LoRA module for supervised fine-tuning, demonstrating their contributions to  human-aligned assessment. The comparison between Row 3 and Row 5 confirms that skill optimization can further enhance model performance by providing more effective evaluation guidance. Meanwhile, Row 4 reveals that skill optimization alone cannot replace parameter fine-tuning, suggesting that the combination of skill optimization and supervised fine-tuning is essential for achieving strong human-aligned evaluation capability. Finally, the results from Rows 5–8 demonstrate that the selected backbone provides better evaluation ability compared with other large multimodal model backbones.

\subsection{Zero-shot Results on Other Benchmarks}
Table~\ref{other_dataset} presents the performance of MIEScore compared with other advanced MLLMs and IEQA-specialized MLLMs on several single-source TIE benchmarks. MIEScore is adapted to the single-source setting by slightly modifying the evaluation prompt while keeping the model parameters unchanged. For benchmarks with pairwise annotations, including GenAI-Bench~\cite{genai}, AROURA-Bench~\cite{aurora}, and EditScore-Bench~\cite{editscore}, we report prediction accuracy. For benchmarks with pointwise annotations, including ImagenHub~\cite{imagenhub}, EBench-18K~\cite{lmm4edit}, EditReward-Bench~\cite{editreward} and IEQA \cite{IEQA}, we report SRCC with human ratings. For benchmarks that provide multiple evaluation dimensions, including ImagenHub, EBench-18K, EditScore-Bench, EditReward-Bench, and IEQA, we report the average result across different dimensions. The results further validate that MIEScore achieves strong generalization ability, performing well even in single-source image editing evaluation scenarios.

\section{Conclusion}
In this paper, we introduce MIE-Bench, the first large-scale benchmark for multi-source image editing evaluation from multiple dimensions, comprising 36K edited images and 108K MOS annotations. Based on MIE-Bench, we propose MIEScore, the first human-aligned evaluation model that jointly assesses visual quality, instruction following, and attribute preservation. Extensive experiments demonstrate that MIEScore consistently outperforms existing methods, achieving superior alignment with human preferences and strong generalization ability across diverse image editing datasets.

\clearpage

\twocolumn[
\begin{@twocolumnfalse}
\begin{center}
{\LARGE\bf Supplementary Material for MIEScore}
\end{center}
\vspace{0.2cm}
\end{@twocolumnfalse}
]

\section{Overview}
This supplementary material provides additional information on the data collection, methodology, experiments, and results discussed in the main paper. Section 2 describes the 16 distinct tasks involved in the prompt and image collection process, while Section 3 presents an overview of the 12 image editing models. Section 4 elaborates on the manual annotation process, including the annotation standards, interface design, and management strategy. Section 5 provides an in-depth analysis of the MIE-Bench. Section 6 details the training configuration of MIEScore. Finally, Section 7 presents details of evaluation models and more comparison results.

\section{Editing Tasks Define}
\label{2}
\begin{figure*}
    \centering
    \includegraphics[width=0.95\linewidth]{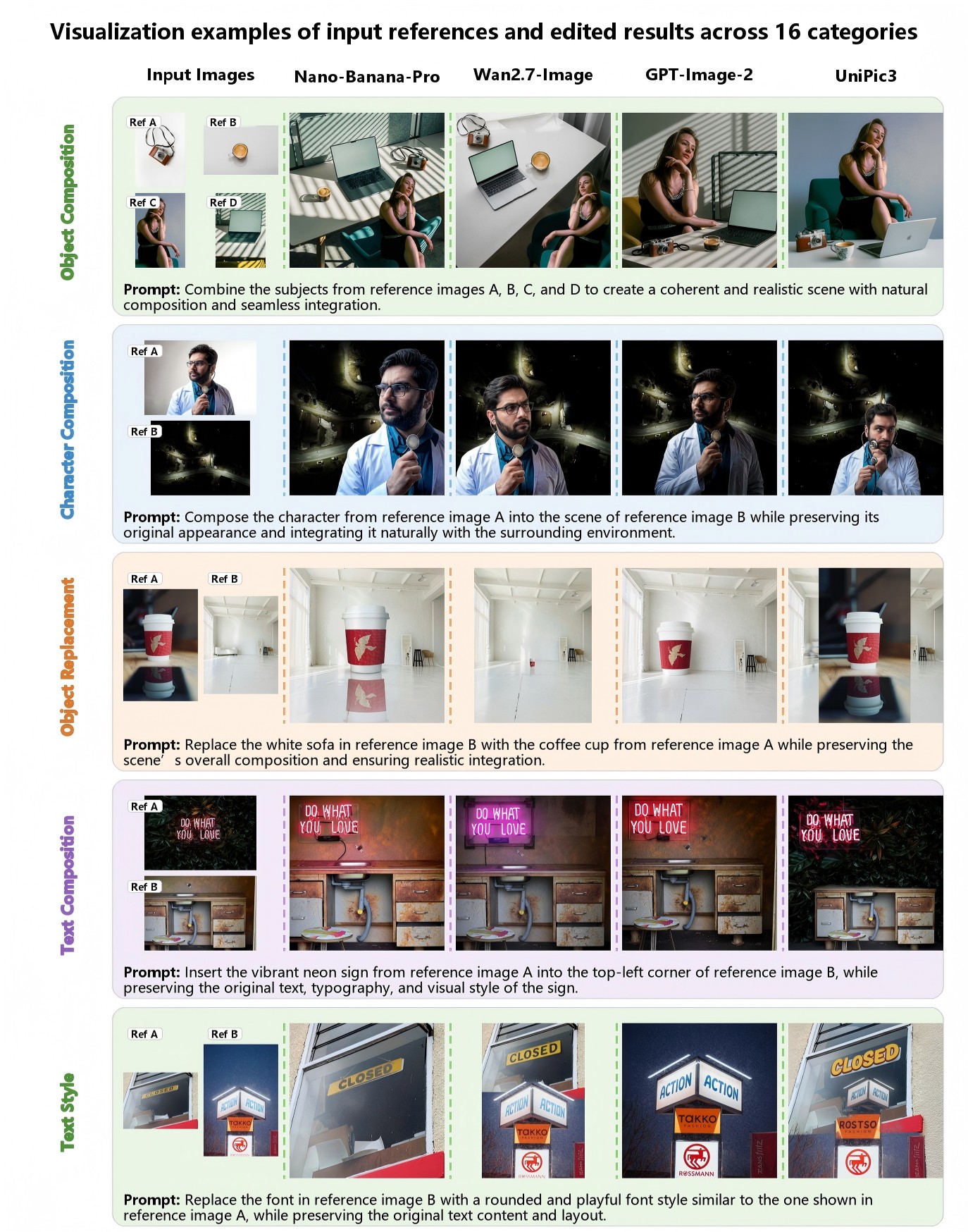}
    \caption{Visualization of different tasks in our MIE-Bench.}
    \label{fig:placeholder1}
\end{figure*}

\begin{figure*}
    \centering
    \includegraphics[width=0.95\linewidth]{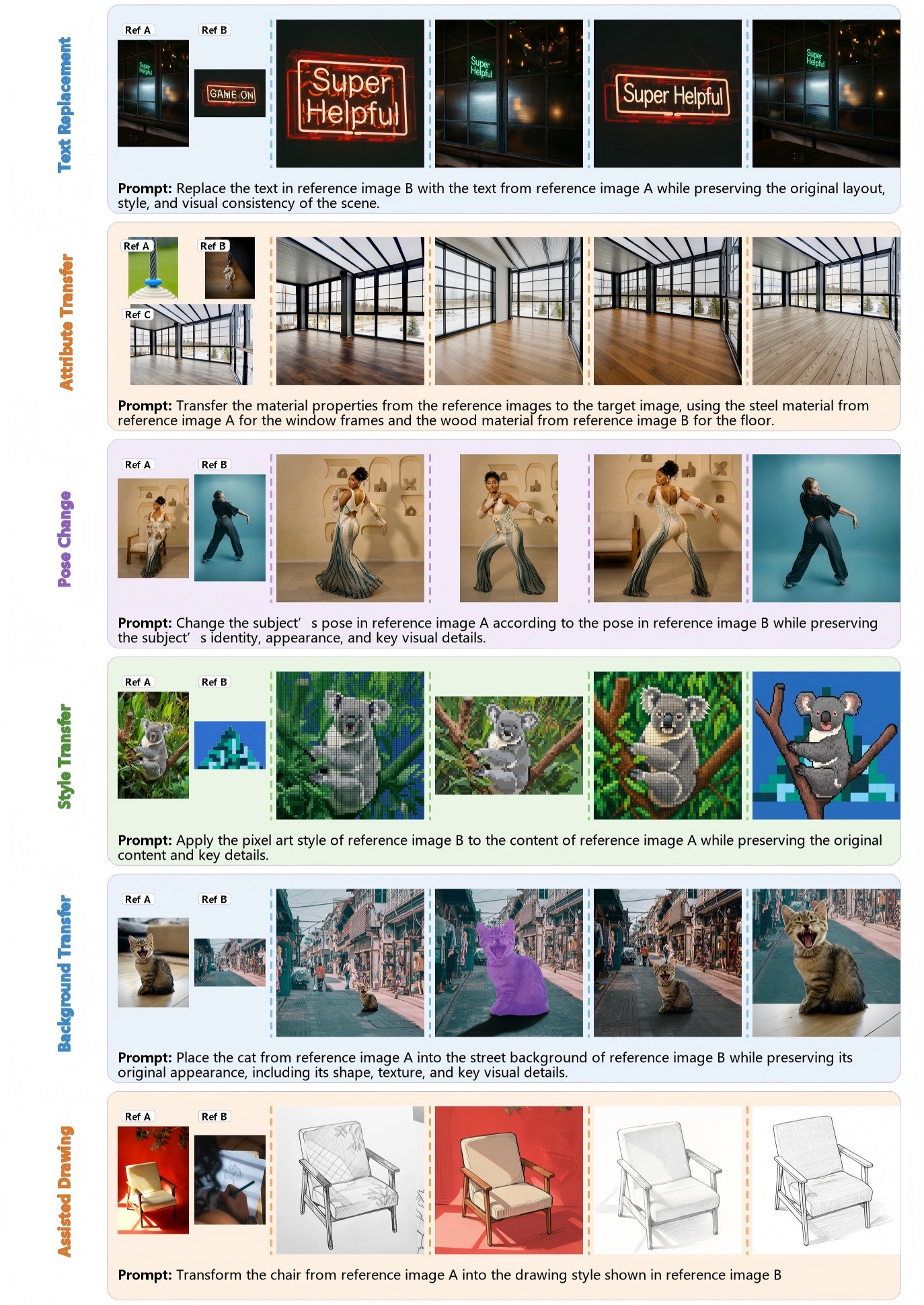}
    \caption{Visualization of different tasks in our MIE-Bench.}
    \label{fig:placeholder2}
\end{figure*}

\begin{figure*}
    \centering
    \includegraphics[width=0.95\linewidth]{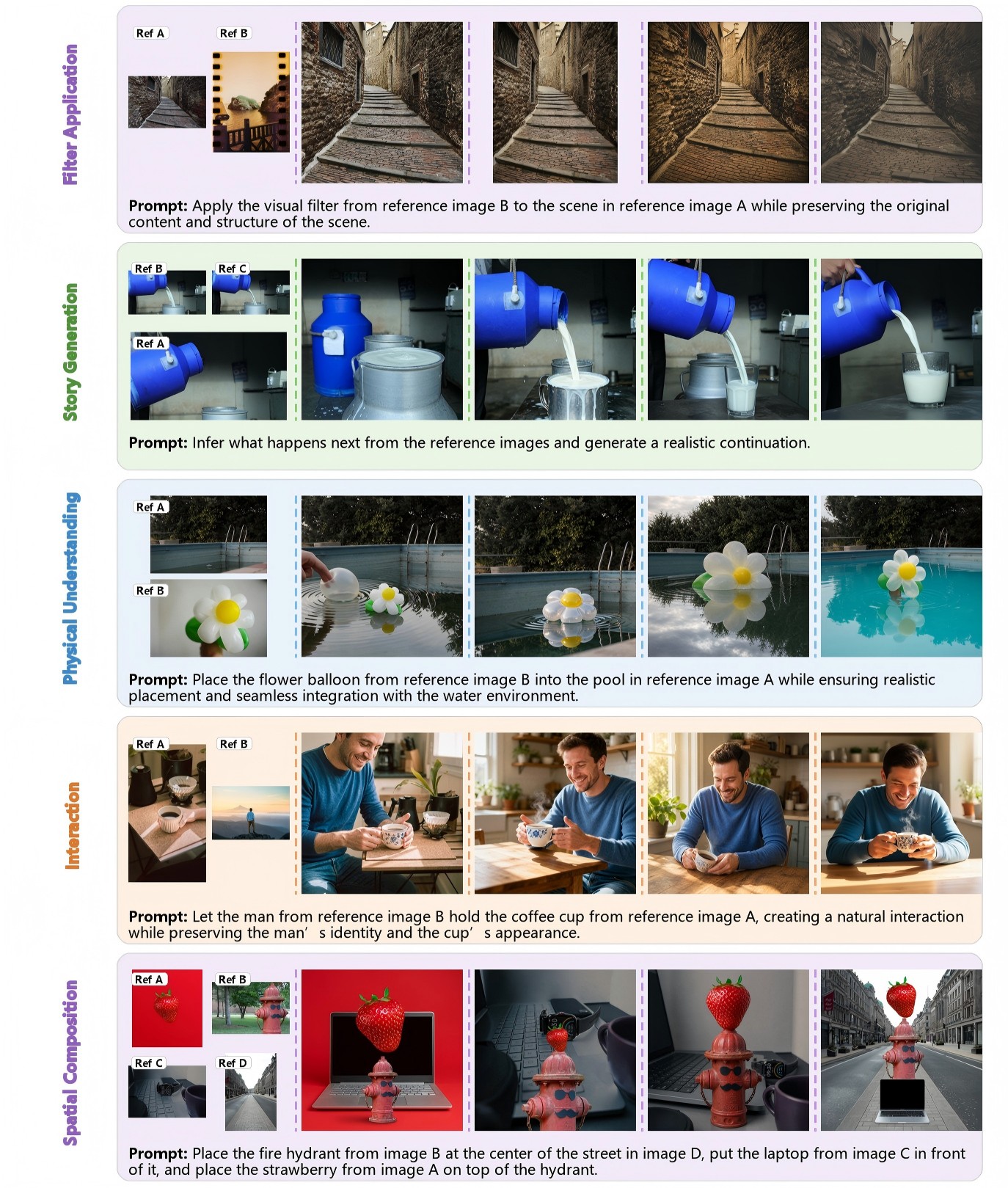}
    \caption{Visualization of different tasks in our MIE-Bench.}
    \label{fig:placeholder3}
\end{figure*}

In this study, we conducted a comprehensive investigation into various categories of multi-source image editing, which we systematically divided into four main dimensions: regional editing, global editing, and complex reasoning editing. Across these dimensions, a total of 16 distinct editing types were defined to capture diverse editing behaviors. Examples are shown in Figure~\ref{fig:placeholder1}-~\ref{fig:placeholder3}.

\textbf{Regional Editing} refers to localized modifications that alter specific regions or entities while preserving the overall image structure. It includes:
\begin{itemize}
\item \textbf{Object Composition}: Combining objects from multiple source images and seamlessly integrating them into a target scene with consistent appearance, lighting, and geometry.

\item \textbf{Character Composition}: Composing human or character subjects from different source images into a unified scene while maintaining identity, appearance, and contextual consistency.

\item \textbf{Object Replacement}: Replacing an existing object with a target object from another source image while preserving the surrounding environment and spatial coherence.

\item \textbf{Text Composition}: Adding or inserting textual elements from source images into a target image with correct content, position, and visual integration.

\item \textbf{Text Style}: Modifying the visual style of text, such as font, color, texture, or artistic appearance, based on reference images.

\item \textbf{Text Replacement}: Replacing existing text with new text content while preserving the original layout, perspective, and style.

\item \textbf{Attribute Transfer}: Transferring visual attributes, such as color, material, texture, pattern, or appearance, from one source image to another target region.

\item \textbf{Pose Change}: Modifying the pose or configuration of a subject according to a reference image while preserving identity and realistic structure.
\end{itemize}

\textbf{Global Editing} refers to image-level transformations that modify the overall visual appearance, environment, or artistic characteristics of an image. It includes:
\begin{itemize}
\item \textbf{Style Transfer}: Transferring global artistic styles, such as painting styles, rendering styles, or visual aesthetics, from reference images to the target image.

\item \textbf{Background Transfer}: Replacing or transforming the background environment using information from reference images while maintaining foreground-background consistency.

\item \textbf{Assisted Drawing}: Completing, extending, or enhancing an image based on sketches, partial drawings, or visual guidance from reference images.

\item \textbf{Filter Composition}: Applying or combining global visual effects, including color grading, illumination adjustment, and photographic filters, according to reference images.
\end{itemize}

\textbf{Complex Reasoning Editing} refers to editing tasks requiring high-level semantic understanding, multi-image reasoning, and physical or contextual knowledge beyond direct visual transfer. It includes:
\begin{itemize}
\item \textbf{Spatial Composition}: Reasoning about spatial relationships among multiple entities and composing objects according to specified positions, depth, scale, and scene structure.

\item \textbf{Interaction}: Understanding and generating meaningful interactions between multiple entities, including human-object, object-object, or character-character relationships.

\item \textbf{Physical Understanding}: Applying knowledge of physical constraints, such as lighting, shadows, reflections, materials, gravity, and object affordances, to produce realistic edits.

\item \textbf{Story Generation}: Creating semantically coherent scenes by combining multiple source images according to high-level narratives, events, or contextual descriptions.
\end{itemize}

% 
% Having the supplementary compiled together with the main paper means that:
% 
% \begin{itemize}
% \item The supplementary can back-reference sections of the main paper, for example, we can refer to \cref{sec:intro};
% \item The main paper can forward reference sub-sections within the supplementary explicitly (e.g. referring to a particular experiment); 
% \item When submitted to arXiv, the supplementary will already included at the end of the paper.
% \end{itemize}

\section{Details of Editing Methods}
\label{3}
\begin{table*}[!t]
\setlength{\tabcolsep}{8pt} 
\centering
\caption{Multi-source image editing models in our MIE-Bench.}
 \resizebox{1\textwidth}{!}{
\begin{tabular}{lcccccccccc}
\toprule
\noalign{\vspace{-1.5pt}}
\textbf{Models}& \textbf{Released} & \textbf{Open Source}& \textbf{URL} \\
 \midrule

BAGEL \cite{bagel}&2025.05&\textcolor{darkgreen}{\ding{51}}& \url{https://github.com/ByteDance-Seed/Bagel} \\
OmniGen2 \cite{omni2}&2025.06&\textcolor{darkgreen}{\ding{51}}&\url{https://github.com/VectorSpaceLab/OmniGen2} \\
DreamOmni2 \cite{dreamomni2}&2025.10&\textcolor{darkgreen}{\ding{51}}& \url{https://github.com/JIA-Lab-research/DreamOmni2} \\
Qwen-Image-Edit \cite{qwenedit}&2025.12&\textcolor{darkgreen}{\ding{51}}& \url{https://github.com/QwenLM/Qwen-Image} \\
UniPic3 \cite{wei2026skyworkunipic30unified}&2026.01&\textcolor{darkgreen}{\ding{51}}&\url{https://github.com/SkyworkAI/UniPic}\\
UniCom \cite{zhao2026unicomunifiedmultimodalmodeling}&2026.03&\textcolor{darkgreen}{\ding{51}}&\url{https://github.com/Tencent-Hunyuan/UniCom}\\
HiDream \cite{hidreamolimage}&2026.05&\textcolor{darkgreen}{\ding{51}}&\url{https://github.com/HiDream-ai/HiDream-I1}\\

 \midrule

 Seedream4 \cite{seedream4}&2025.09&\color{red}{\ding{55}}&\url{https://seed.bytedance.com/zh/seedream4_0}\\
Nano-Banana-Pro \cite{google_nanobananapro}&2025.11&\color{red}{\ding{55}}&\url{https://www.nanobananapro.org/}\\
Seedream5 \cite{seedream5}&2026.02&\color{red}{\ding{55}}&\url{https://seed.bytedance.com/en/seedream5_0_lite}\\
GPT-Image-2 \cite{openai_gptimage2}&2026.04&\color{red}{\ding{55}}&\url{https://chatgpt.com/images/}\\
Wan2.7-Image \cite{mao2026wanimage}&2026.04&\color{red}{\ding{55}}&\url{https://wan2-7.io/image/wan2-7-image/}\\

\bottomrule
\label{MIE models}
\end{tabular}}
\end{table*}

\begin{itemize}

\item \textbf{BAGEL \cite{bagel}} 
is an open-source, decoder-only foundational model that unifies multimodal understanding and generation. Pretrained on trillions of interleaved text, image, video, and web tokens, it demonstrates strong multimodal reasoning in both generation and understanding tasks.

\item \textbf{OmniGen2 \cite{omnigen2}} 
is an instruction-guided image editing model using a Diffusion Transformer (DiT) backbone. It unifies semantic reasoning and generation, supporting complex global edits driven by natural language instructions.

\item \textbf{DreamOmni2 \cite{dreamomni2}}
is a unified model for image generation and editing that jointly learns text-to-image generation and various editing tasks. Leveraging a synthetic data pipeline for high-quality instruction- and drag-based editing data.

\item \textbf{Qwen-Image-Edit \cite{qwenedit}} 
is an instruction-based image editing system aligned with the Qwen multi-modal model family. Built on a DiT backbone, it interprets high-level natural language instructions for semantically consistent global edits.

\item \textbf{UniPic3 \cite{wei2026skyworkunipic30unified}} is a unified multimodal framework that supports both single-image editing and multi-image composition, handling 1–6 input images with arbitrary resolutions through a sequence-modeling-based generation paradigm. By introducing a high-quality data pipeline and efficient post-training acceleration, it achieves state-of-the-art performance on image editing and multi-image composition benchmarks.

\item \textbf{UniCom \cite{zhao2026unicomunifiedmultimodalmodeling}} introduces a unified multimodal framework that bridges visual understanding and generation by using compressed continuous representations instead of discrete visual tokens. With an attention-based semantic compressor and a transfusion architecture, it preserves rich semantic information, improves convergence and consistency, and achieves state-of-the-art performance in unified generation and controllable image editing without relying on VAE.

\item \textbf{HiDream \cite{hidreamolimage}} is a 17B-parameter open-source image generation foundation model that introduces a sparse Diffusion Transformer with dynamic Mixture-of-Experts (MoE) to achieve high-quality generation with improved efficiency. It further extends to instruction-based image editing and an interactive image creation agent, providing a unified framework for generation, editing, and refinement.

\item \textbf{SeedDream4 \cite{seedream4}} is an advanced image generation and editing model designed for high-quality visual creation with strong semantic understanding, supporting both text-to-image generation and multi-image composition tasks. It achieves superior consistency, detail preservation, and controllable editing performance by leveraging large-scale multimodal training and advanced generative modeling techniques, making it a strong foundation model for complex image creation workflows.

\item \textbf{Nano-Banana-Pro \cite{nanobanana}} is a powerful multimodal image generation and editing model that emphasizes high-fidelity image synthesis, precise instruction following, and multi-image composition. It demonstrates strong capabilities in maintaining subject consistency, integrating multiple visual references, and performing complex image transformations, making it effective for advanced creative editing and real-world visual generation tasks.

\item \textbf{SeedDream5 \cite{seedream5}} is a next-generation multimodal image generation model that unifies text-to-image generation, image editing, and multi-image composition within a single framework. It improves visual reasoning, instruction following, and consistency through advanced multimodal modeling, enabling high-fidelity generation, precise editing, and complex reference-based image creation.

\item \textbf{GPT-Image-2 \cite{openai_gptimage2}} is OpenAI’s advanced image generation and editing model that combines strong visual understanding with high-fidelity image synthesis, enabling precise instruction following, realistic generation, and complex image editing. It significantly improves text rendering, detail preservation, and multi-step visual reasoning, supporting controllable creation and refinement across diverse image generation scenarios.

\item \textbf{Wan2.7-Image \cite{mao2026wanimage}} is a unified multimodal image generation and editing model that supports high-quality image synthesis, instruction-based editing, and multi-image composition. It leverages advanced multimodal understanding to achieve strong semantic alignment, visual consistency, and controllable generation, enabling complex image creation and manipulation across diverse scenarios.

\end{itemize}

\section{Details of Manual Annotation Process}
\label{4}
\subsection{Annotation Dimension and Criteria}
To comprehensively evaluate the performance of edited images, we propose a triple-dimensional evaluation framework that assesses visual quality, instruction following and attribute preservation. This framework enables a thorough analysis of various aspects of multi-source image editing, offering a comprehensive understanding of a model’s strengths and limitations.
\begin{itemize}
\item \textbf{visual quality:} This dimension focuses on the overall visual quality of the edited image from a human perception perspective. It considers multiple aspects such as authenticity (i.e., whether the image looks realistic and natural), the absence of artifacts or distortions, accurate and harmonious color rendering, and the richness and clarity of visual details. A high visual quality image should be visually pleasing and indistinguishable from a naturally captured one.

\item \textbf{instruction following:} This dimension evaluates how precisely the content of the edited image aligns with the given editing instruction or textual prompt. It examines whether the intended changes have been correctly and completely applied, without over- or under-editing. The evaluation emphasizes semantic accuracy and the faithful execution of the editing intent, ensuring the image reflects the described transformation or manipulation.

\item \textbf{Attribute Preservation:} This dimension assesses how well the edited image maintains the original visual attributes of the unedited regions in the source images. It focuses on preserving the context, structure, and appearance of unchanged areas, ensuring they are not unintentionally modified or degraded. This includes maintaining consistent lighting, texture, identity (in the case of humans or objects), and spatial relationships, contributing to the overall coherence and realism of the final result.
\end{itemize}

\begin{figure*}
    \centering
    \includegraphics[width=\linewidth]{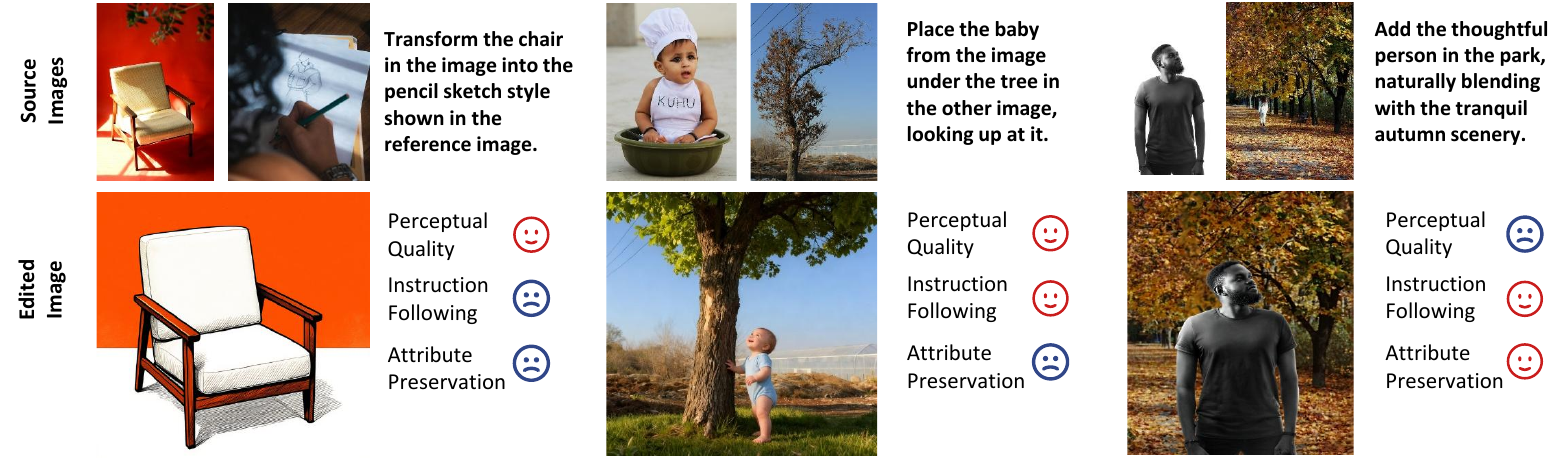}
    \caption{Illustration of the evaluation dimensions of image editing: visual quality, instruction following and attribute preservation, attached with examples with different subjective qualities.}
    \label{3d}
\end{figure*}

Evaluating image editing quality through visual quality, instruction following, and attribute preservation is essential for a comprehensive assessment. visual quality ensures that the edited image is visually appealing, realistic, and free from artifacts or distortions, which is crucial for meeting aesthetic expectations. Without it, an image may appear unnatural, even if the editing process followed the instructions. instruction following guarantees that the changes made to the image match the given instructions or textual prompts. If this dimension is neglected, the final image may fail to reflect the intended edits, even if the image has good visual quality and attribute preservation. Attribute preservation ensures that the unedited parts of the image retain their original attributes, such as lighting, texture, and structure. If ignored, the image may become inconsistent with the source image, despite completing the editing instructions. These three dimensions are interdependent and complement each other, addressing different aspects of image quality. Omitting any one of them would result in an incomplete evaluation, potentially leading to an image that is aesthetically pleasing but not accurate or coherent, ultimately failing to meet the user's expectations. Figure~\ref{3d} provides examples illustrating the importance of these three evaluation dimensions.

\subsection{Subjective Experiment of Scoring}
\begin{figure*}
    \centering
    \includegraphics[width=\linewidth]{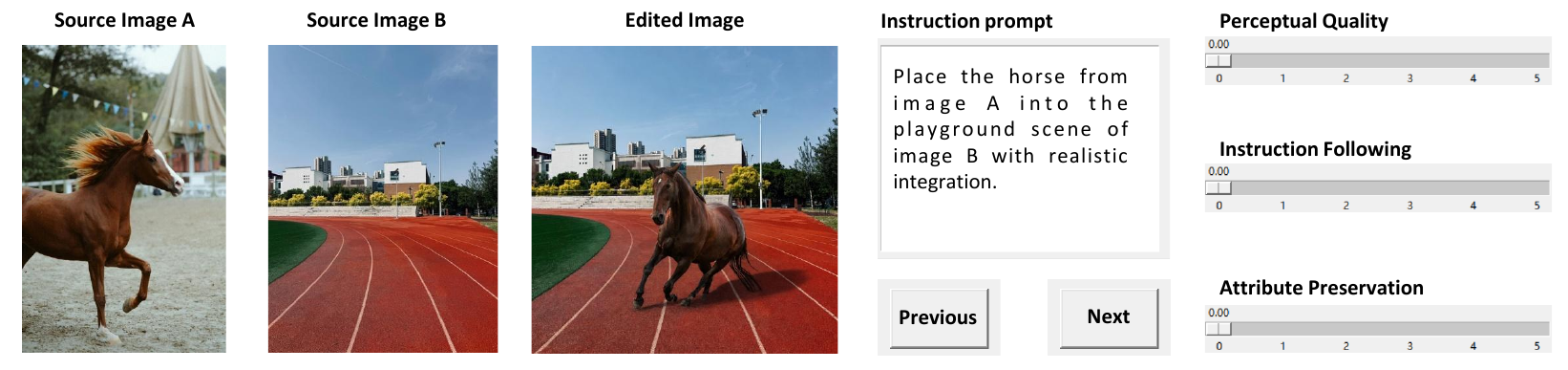}
    \caption{An example of the annotation interface for subjective experiment of Scoring.}
    \label{gui}
\end{figure*}

We have developed a custom annotation interface for rating, shown in Figure~\ref{gui}. This manual evaluation platform is built using the Python tkinter package and is designed to facilitate MOS assessments. In this experiment, participants evaluate images based on three independent dimensions related to a specific task challenge.

Each trial presents instruction editing prompt, one source image and one edited image randomly selected from 23 different models. Participants are instructed to assign absolute scores to each image based on the three predefined dimensions, rather than making relative comparisons. For each image, participants provide three separate scores representing the three evaluation dimensions. The scoring annotation task involved 5 participants rating each image on a 0-5 Likert scale, assessing visual quality, instruction following and attribute preservation. 

Before engaging in the annotation tasks, all participants undergo a rigorous training process. They received detailed instructions and multiple standardized examples. The specific rating instructions are given below:

\textbf{Visual Quality:}
\begin{itemize}
\item \textbf{Score 0-1:} The image exhibits severe blurriness, visible artifacts, and significant color distortion, which drastically deviate from logical realism. These issues make the image difficult to recognize, resulting in an overall unattractive appearance. The flaws are so prominent that the image appears unrealistic, making it unsuitable for viewing in any context. The image feels fundamentally damaged, and its content is largely incomprehensible.
\item \textbf{Score 1-2:} The image has substantial issues with clarity, visible artifacts, and color inaccuracies, significantly affecting its overall quality. Although the image remains somewhat recognizable, the defects make it difficult to appreciate. The distortion and artifacts hinder proper understanding, but the content is still somewhat discernible. The flaws are considerable, but the image can still be viewed with effort.
\item \textbf{Score 2-3:} The image has notable flaws in clarity, visible artifacts, and issues with color accuracy and logical realism. While these problems are significant, the general content of the image remains recognizable. However, the overall visual quality is compromised, and the viewer may find it difficult to fully appreciate the image due to these issues. The flaws affect the image but do not render it entirely unrecognizable. The image is still legible, but its visual appeal is diminished.
\item \textbf{Score 3-4:} The image has minor flaws in clarity, artifacts, color accuracy, and logical realism, which only slightly affect the viewer's visual experience. While the issues are noticeable, they do not severely detract from the overall quality of the image. The content remains mostly clear and coherent, and the image is still pleasant to look at, but the minor imperfections may cause slight discomfort upon closer inspection. The visual impact is generally positive, with only small disruptions in quality.
\item \textbf{Score 4-5:} The image has high overall quality, with clear and easily recognizable visuals. The details are rich, and the image maintains logical realism throughout, making it highly believable. There are no significant flaws or distractions, and the image presents itself in a way that is visually pleasing and accurately reflects its intended purpose or scene. The image stands out for its clarity and aesthetic appeal, making it easy to interpret and highly enjoyable to view.
\end{itemize}

\textbf{instruction following:}
\begin{itemize}
\item \textbf{Score 0-1:} No editing was applied, or the edits made are completely inconsistent with the expectations set by the instructions. The image does not resemble the requested outcome at all, making it unrecognizable or irrelevant to the task. The changes are so far from what was expected that the image fails to fulfill its intended purpose.
\item \textbf{Score 1-2:} Some editing was applied, but there are substantial deviations from the expected outcome described in the instructions. The changes made are noticeable but fail to capture the intended transformation accurately. The result is significantly different from what was expected, though some elements may still bear a faint resemblance to the editing instructions.
\item \textbf{Score 2-3:} Editing was applied, but there are significant deviations from the expected outcome described in the textual instructions. The changes made do not align well with the intended transformation, and the result is noticeably different from what was expected, although the general idea may still be somewhat recognizable.
\item \textbf{Score 3-4:} The editing applied to the image but partially deviates from the intended result. There are some discrepancies between the expected and actual edits, but the overall changes are still somewhat aligned with the editing instructions, with only minor inconsistencies.
\item \textbf{Score 4-5:} The editing fully meets the expectations outlined in the textual instructions. All changes, such as color, shape, position, and effects, have been applied accurately and completely, and the final image is a precise reflection of the intended transformation, without any noticeable discrepancies.
\end{itemize}

\textbf{Attribute Preservation:}
\begin{itemize}
\item \textbf{Score 0-1:} The edited image is largely unrelated to all source images. The generated result fails to preserve the key information, attributes, or identities from the references, and most source images are ignored or incorrectly represented.

\item \textbf{Score 1-2:} The edited image shows severe inconsistencies with the source images. It only follows one or a few references while ignoring other provided images, or introduces substantial unrelated content. Important attributes, structures, or details from the original sources are incorrectly preserved.

\item \textbf{Score 2-3:} The edited image incorporates information from some source images but fails to fully consider all references. Some key elements, attributes, or styles from one or more source images are missing, distorted, or inconsistent, resulting in incomplete multi-source fusion.

\item \textbf{Score 3-4:} The edited image successfully incorporates information from all source images, with only minor inconsistencies in subject attributes, style, details, or relationships. These deviations do not significantly affect the overall understanding of the intended multi-source editing result.

\item \textbf{Score 4-5:} The edited image correctly and reasonably integrates information from all source images. The key attributes, identities, structures, and styles of each reference are faithfully preserved, achieving coherent and seamless multi-source fusion.
\end{itemize}

Moreover, to ensure a high level of understanding and consistency, a pre-test is conducted to assess participants’ comprehension of the criteria and their alignment with the standard examples. Participants who do not meet the required accuracy threshold are excluded from further participation, ensuring that only well-prepared individuals contributed to the final dataset. 

During the experiment, all evaluations are conducted in a controlled laboratory environment under normal indoor lighting conditions. Participants are seated at a comfortable viewing distance of approximately 60 cm from the screen to minimize visual strain and ensure consistent evaluation conditions. The whole process cost 650 working hours.

\subsection{Ethics Statement}
To ensure ethical compliance and maintain high annotation quality, we implemented a comprehensive process for the MIE-Bench. All participants were fully informed about the experiment’s objectives, tasks, and ethical considerations. Each participant signed an informed consent agreement, granting permission for their annotations to be used exclusively for noncommercial research purposes. The dataset, consisting edited images along with their editing prompts and source images, has been publicly released under the CC BY 4.0 license, ensuring accessibility while adhering to ethical guidelines. We implemented a rigorous manual review process during the image generation phase to ensure the exclusion of any inappropriate or NSFW content (both textual and visual). This step guaranteed that the dataset remained suitable for academic and research purposes. 

This rigorous and ethically sound annotation management strategy establishes MIE-Bench as a robust and reliable resource for advancing research in image quality assessment.
\begin{table*}[!t]
  \setlength{\tabcolsep}{8pt}
\renewcommand{\arraystretch}{0.85}
\centering
\caption{Performance comparisons of editing models on human annotation visual quality score.}
\label{cvisual}
 \resizebox{1\textwidth}{!}{
\begin{tabular}{lcccccccccccccccc}
\toprule
&\multicolumn{8}{c}{Regional Editing}\\
\cmidrule(lr){2-9}
Model/Task & \makecell{Attribute\\Transfer} & \makecell{Character\\Composition} & \makecell{Object\\Composition} & \makecell{Object\\Replacement} & \makecell{Pose\\Change} & \makecell{Text\\Composition} & \makecell{Text\\Style} & \makecell{Text\\Replacement} \\
\midrule
DreamOmni2 \cite{dreamomni2} & \mbluebf{54.70} & 42.12 & 43.98 & 52.16 & 55.75 & 50.12 & \mbluebf{58.22} & 55.40 \\
Nano-Banana-Pro \cite{google_nanobananapro} & 50.67 & 47.85 & 50.55 & 46.85 & 56.50 & 48.40 & 58.12 & \mbluebf{61.57} \\
OmniGen2 \cite{omni2} & 38.80 & 52.74 & 48.37 & 37.80 & 54.51 & 44.49 & 42.31 & 37.76 \\
Qwen-Image-Edit \cite{qwenedit} & 48.93 & 43.69 & 43.84 & 49.01 & 45.08 & 44.73 & 40.04 & 49.37 \\
Seedream4 \cite{seedream4} & 53.30 & \mbluebf{55.81} & \mredbf{61.26} & \mbluebf{55.51} & \mbluebf{57.68} & \mbluebf{56.09} & 57.24 & 59.65 \\
Seedream5 \cite{seedream5} & 50.39 & 51.11 & \mbluebf{57.96} & 52.06 & 55.42 & 54.26 & 56.20 & 57.80 \\
UniPic3 \cite{wei2026skyworkunipic30unified} & 50.90 & 44.97 & 51.79 & 53.93 & 51.79 & 47.35 & 33.42 & 41.22 \\
BAGEL \cite{bagel} & 26.15 & 32.95 & 30.66 & 24.33 & 39.47 & 33.33 & 27.92 & 40.31 \\
GPT-Image-2 \cite{openai_gptimage2} & \mredbf{55.88} & \mredbf{56.95} & 56.64 & \mredbf{56.03} & 57.18 & \mredbf{59.13} & \mredbf{62.25} & \mredbf{63.19} \\
HiDream \cite{hidreamolimage} & 49.08 & 51.06 & 52.48 & 50.37 & 43.86 & 46.82 & 46.43 & 48.02 \\
UniCom \cite{zhao2026unicomunifiedmultimodalmodeling} & 47.18 & 50.07 & 54.47 & 52.46 & 53.67 & 51.71 & 49.23 & 29.58 \\
Wan2.7-Image \cite{mao2026wanimage} & 53.38 & 50.50 & 51.98 & 50.44 & \mredbf{59.55} & 54.38 & 54.78 & 59.52 \\
\bottomrule
\end{tabular}}
 \resizebox{1\textwidth}{!}{
\begin{tabular}{lcccccccccccccccc}
\toprule
&\multicolumn{4}{c}{Global Editing}&\multicolumn{4}{c}{Complex Reasoning Editing}\\
\cmidrule(lr){2-5}
\cmidrule(lr){6-9}
Model/Task & \makecell{Background\\Transfer} & \makecell{Filter\\Composition} & \makecell{Style\\Transfer} & \makecell{Assisted\\Drawing} & \makecell{Story\\Generation} & Interaction & \makecell{Physical\\Understanding} & \makecell{Spatial\\Composition} \\
\midrule
DreamOmni2 \cite{dreamomni2} & \mredbf{55.62} & 47.24 & 53.52 & 46.28 & 43.61 & 48.83 & 45.91 & 46.19 \\
Nano-Banana-Pro \cite{google_nanobananapro} & 41.65 & 53.73 & 56.35 & 55.69 & 58.23 & 61.39 & 49.54 & 47.63 \\
OmniGen2 \cite{omni2} & 33.62 & 53.50 & 54.40 & 56.59 & 50.26 & 53.15 & 50.63 & 44.77 \\
Qwen-Image-Edit \cite{qwenedit} & 40.11 & 38.76 & 48.04 & 49.29 & 43.34 & 51.51 & 45.58 & 42.86 \\
Seedream4 \cite{seedream4} & \mbluebf{50.78} & 50.17 & \mbluebf{59.04} & 53.73 & 58.10 & \mredbf{65.54} & \mbluebf{54.29} & \mredbf{55.24} \\
Seedream5 \cite{seedream5} & 49.09 & 50.81 & 54.13 & 50.97 & 54.97 & 62.03 & 53.11 & 49.30 \\
UniPic3 \cite{wei2026skyworkunipic30unified} & 40.98 & \mbluebf{53.94} & 53.91 & \mredbf{56.87} & 52.34 & 63.30 & 50.19 & 49.17 \\
BAGEL \cite{bagel} & 3.87 & 34.30 & 28.97 & 46.63 & 49.25 & 59.62 & 43.68 & 27.71 \\
GPT-Image-2 \cite{openai_gptimage2} & 45.69 & \mredbf{56.55} & \mredbf{61.77} & \mbluebf{56.75} & \mbluebf{60.04} & \mbluebf{65.38} & \mredbf{59.44} & \mbluebf{54.98} \\
HiDream \cite{hidreamolimage} & 17.35 & 41.67 & 48.70 & 51.08 & 51.65 & 62.73 & 50.71 & 44.24 \\
UniCom \cite{zhao2026unicomunifiedmultimodalmodeling} & 45.49 & 38.86 & 56.67 & 47.86 & 55.10 & 59.56 & 52.83 & 48.50 \\
Wan2.7-Image \cite{mao2026wanimage} & 29.24 & 52.50 & 57.19 & 56.37 & \mredbf{60.60} & 63.11 & 53.94 & 50.48 \\
\bottomrule
\end{tabular}}
\end{table*}

\begin{table*}[!t]
  \setlength{\tabcolsep}{8pt}
\renewcommand{\arraystretch}{0.85}
\centering
\caption{Performance comparisons of editing models on human annotation instruction following score.}
\label{cediting}
 \resizebox{1\textwidth}{!}{
\begin{tabular}{l||cccccccccccccccccccccc}
\toprule
&\multicolumn{8}{c}{Regional Editing}\\
\cmidrule(lr){2-9}
Model/Task & \makecell{Attribute\\Transfer} & \makecell{Character\\Composition} & \makecell{Object\\Composition} & \makecell{Object\\Replacement} & \makecell{Pose\\Change} & \makecell{Text\\Composition} & \makecell{Text\\Style} & \makecell{Text\\Replacement} \\
\midrule
DreamOmni2 \cite{dreamomni2} & 38.05 & 45.07 & 43.21 & 33.72 & 23.49 & 36.33 & 21.26 & 31.99 \\
Nano-Banana-Pro \cite{google_nanobananapro} & \mredbf{61.97} & \mbluebf{63.75} & \mredbf{64.59} & \mredbf{60.25} & \mbluebf{48.72} & \mbluebf{61.98} & \mredbf{57.44} & \mbluebf{59.04} \\
OmniGen2 \cite{omni2} & 53.27 & 59.96 & 53.38 & 49.65 & 31.98 & 51.55 & 33.07 & 41.35 \\
Qwen-Image-Edit \cite{qwenedit} & 56.45 & 60.31 & 57.80 & 51.66 & 34.51 & 56.57 & 37.66 & 44.30 \\
Seedream4 \cite{seedream4} & 51.78 & 59.04 & 62.82 & 42.05 & 45.32 & 56.77 & 39.56 & 50.36 \\
Seedream5 \cite{seedream5} & 54.22 & 59.83 & \mbluebf{63.00} & 44.94 & 46.92 & 56.73 & 37.79 & 49.54 \\
UniPic3 \cite{wei2026skyworkunipic30unified} & 55.85 & 59.16 & 60.64 & 45.71 & 33.26 & 56.94 & 40.76 & 45.52 \\
BAGEL \cite{bagel} & 55.66 & 58.46 & 54.98 & 45.92 & 33.92 & 50.34 & 37.78 & 32.57 \\
GPT-Image-2 \cite{openai_gptimage2} & \mbluebf{61.76} & \mredbf{64.02} & 61.53 & \mbluebf{58.72} & \mredbf{52.97} & \mredbf{64.28} & \mbluebf{56.32} & \mredbf{59.69} \\
HiDream \cite{hidreamolimage} & 57.23 & 59.34 & 57.33 & 46.50 & 35.49 & 58.31 & 39.69 & 44.80 \\
UniCom \cite{zhao2026unicomunifiedmultimodalmodeling} & 50.76 & 57.08 & 57.38 & 43.34 & 22.97 & 55.36 & 41.66 & 37.14 \\
Wan2.7-Image \cite{mao2026wanimage} & 56.77 & 61.72 & 61.81 & 52.39 & 44.97 & 61.24 & 50.23 & 48.26 \\
\bottomrule
\end{tabular}}

 \resizebox{1\textwidth}{!}{
\begin{tabular}{l||ccccccccccccccccccccc:c}
\toprule
&\multicolumn{4}{c}{Global Editing}&\multicolumn{4}{c}{Complex Reasoning Editing}\\
\cmidrule(lr){2-5}
\cmidrule(lr){6-9}
Model/Task & \makecell{Background\\Transfer} & \makecell{Filter\\Composition} & \makecell{Style\\Transfer} & \makecell{Assisted\\Drawing} & \makecell{Story\\Generation} & Interaction & \makecell{Physical\\Understanding} & \makecell{Spatial\\Composition}\\
\midrule
DreamOmni2 \cite{dreamomni2} & 42.56 & 16.53 & 49.53 & 30.56 & 39.33 & 38.29 & 17.90 & 51.87 \\
Nano-Banana-Pro \cite{google_nanobananapro} & \mbluebf{56.01} & \mbluebf{59.30} & 59.62 & \mbluebf{62.71} & 53.28 & \mbluebf{62.72} & 42.69 & 59.53 \\
OmniGen2 \cite{omni2} & 40.92 & 40.37 & 43.62 & 50.59 & 42.41 & 55.29 & 15.63 & 54.85 \\
Qwen-Image-Edit \cite{qwenedit} & 53.21 & 33.65 & 51.54 & 53.15 & 42.83 & 58.65 & 12.09 & 52.04 \\
Seedream4 \cite{seedream4} & 53.50 & 20.71 & 59.19 & 59.48 & 52.86 & 60.15 & 11.25 & \mbluebf{62.45} \\
Seedream5 \cite{seedream5} & 52.07 & 19.38 & \mbluebf{60.36} & 60.05 & 51.76 & 57.97 & \mredbf{51.74} & \mredbf{62.74} \\
UniPic3 \cite{wei2026skyworkunipic30unified} & 47.80 & 31.77 & 57.78 & 51.01 & 45.66 & 61.46 & 39.24 & 59.33 \\
BAGEL \cite{bagel} & 41.20 & 47.14 & 49.94 & 36.95 & 44.36 & 55.70 & 41.18 & 56.13 \\
GPT-Image-2 \cite{openai_gptimage2} & \mredbf{56.73} & \mredbf{59.43} & \mredbf{61.87} & \mredbf{63.10} & \mredbf{55.62} & \mredbf{64.11} & \mbluebf{49.02} & 62.17 \\
HiDream \cite{hidreamolimage} & 41.33 & 38.26 & 50.51 & 44.41 & 49.33 & 55.84 & 13.41 & 60.10 \\
UniCom \cite{zhao2026unicomunifiedmultimodalmodeling} & 37.12 & 17.71 & 41.16 & 54.87 & \mbluebf{55.19} & 54.94 & 28.78 & 61.06 \\
Wan2.7-Image \cite{mao2026wanimage} & 52.64 & 52.54 & 60.19 & 59.34 & 48.51 & 60.08 & 39.57 & 59.50 \\
\bottomrule
\end{tabular}}
\end{table*}

\begin{table*}[!t]
  \setlength{\tabcolsep}{8pt}
\renewcommand{\arraystretch}{0.85}
\centering
\caption{Performance comparisons of editing models on human annotation attribute preservation score.}
\label{cpreservation}
 \resizebox{1\textwidth}{!}{
\begin{tabular}{l||cccccccccccccccccccccc}
\toprule
&\multicolumn{8}{c}{Regional Editing}\\
\cmidrule(lr){2-9}
Model/Task & \makecell{Attribute\\Transfer} & \makecell{Character\\Composition} & \makecell{Object\\Composition} & \makecell{Object\\Replacement} & \makecell{Pose\\Change} & \makecell{Text\\Composition} & \makecell{Text\\Style} & \makecell{Text\\Replacement} \\
\midrule
DreamOmni2 \cite{dreamomni2} & 36.43 & 45.24 & 48.57 & 36.03 & 39.94 & 37.91 & 36.27 & 36.82 \\
Nano-Banana-Pro \cite{google_nanobananapro} & 52.41 & 63.17 & 68.28 & \mredbf{67.21} & \mbluebf{54.40} & \mbluebf{59.85} & \mredbf{57.57} & 46.44 \\
OmniGen2 \cite{omni2} & 50.32 & 55.39 & 58.06 & 56.70 & 35.11 & 50.51 & 40.88 & 44.81 \\
Qwen-Image-Edit \cite{qwenedit} & 48.45 & 56.19 & 60.23 & 50.48 & 39.94 & 46.33 & 40.37 & 42.11 \\
Seedream4 \cite{seedream4} & 46.71 & 53.38 & 61.11 & 43.04 & 49.39 & 49.63 & 39.20 & 41.10 \\
Seedream5 \cite{seedream5} & 48.35 & 53.03 & 63.29 & 44.08 & 51.39 & 50.51 & 39.34 & 42.20 \\
UniPic3 \cite{wei2026skyworkunipic30unified} & 51.08 & 55.99 & 66.19 & 44.30 & 40.69 & 51.10 & 44.33 & \mbluebf{47.75} \\
BAGEL \cite{bagel} & 51.42 & 55.23 & 60.84 & 58.12 & 42.30 & 49.16 & 40.53 & 39.60 \\
GPT-Image-2 \cite{openai_gptimage2} & \mredbf{57.47} & \mredbf{66.21} & \mredbf{69.84} & \mbluebf{62.86} & \mredbf{59.37} & \mredbf{62.09} & \mbluebf{57.24} & 46.91 \\
HiDream \cite{hidreamolimage} & \mbluebf{57.04} & 58.51 & 64.05 & 47.42 & 42.19 & 56.82 & 43.99 & \mredbf{48.93} \\
UniCom \cite{zhao2026unicomunifiedmultimodalmodeling} & 35.40 & 40.16 & 48.72 & 39.31 & 16.63 & 42.15 & 26.71 & 23.72 \\
Wan2.7-Image \cite{mao2026wanimage} & 56.71 & \mbluebf{63.66} & \mbluebf{69.34} & 60.31 & 54.19 & 58.17 & 53.11 & 47.04 \\
\bottomrule
\end{tabular}}

 \resizebox{1\textwidth}{!}{
\begin{tabular}{l||ccccccccccccccccccccc:c}
\toprule
&\multicolumn{4}{c}{Global Editing}&\multicolumn{4}{c}{Complex Reasoning Editing}\\
\cmidrule(lr){2-5}
\cmidrule(lr){6-9}
Model/Task & \makecell{Background\\Transfer} & \makecell{Filter\\Composition} & \makecell{Style\\Transfer} & \makecell{Assisted\\Drawing} & \makecell{Story\\Generation} & Interaction & \makecell{Physical\\Understanding} & \makecell{Spatial\\Composition}\\
\midrule
DreamOmni2 \cite{dreamomni2} & 50.67 & 38.95 & 49.54 & 33.95 & 43.67 & 44.34 & 2.58 & 43.41 \\
Nano-Banana-Pro \cite{google_nanobananapro} & \mbluebf{60.67} & \mbluebf{67.71} & 58.70 & \mbluebf{62.79} & \mredbf{56.47} & 61.95 & 44.45 & \mredbf{54.91} \\
OmniGen2 \cite{omni2} & 45.11 & 47.33 & 43.58 & 51.28 & 49.23 & 54.10 & 2.82 & 48.22 \\
Qwen-Image-Edit \cite{qwenedit} & 56.99 & 40.92 & 51.33 & 52.00 & 48.24 & 57.20 & 3.86 & 47.77 \\
Seedream4 \cite{seedream4} & 58.08 & 33.37 & 57.62 & 59.50 & 50.98 & 53.69 & 1.43 & 47.03 \\
Seedream5 \cite{seedream5} & 57.17 & 34.20 & \mbluebf{60.11} & 61.21 & 51.21 & 50.66 & 38.35 & 49.69 \\
UniPic3 \cite{wei2026skyworkunipic30unified} & 52.98 & 48.46 & 57.68 & 51.79 & 53.69 & 60.00 & 39.61 & 49.81 \\
BAGEL \cite{bagel} & 48.92 & 58.06 & 51.87 & 42.22 & 47.72 & 55.65 & \mredbf{56.46} & 50.96 \\
GPT-Image-2 \cite{openai_gptimage2} & \mredbf{63.02} & \mredbf{69.35} & \mredbf{62.17} & \mredbf{67.75} & 51.30 & \mbluebf{63.55} & 44.96 & 51.74 \\
HiDream \cite{hidreamolimage} & 47.51 & 46.36 & 50.28 & 50.29 & 49.81 & 60.47 & 2.55 & 49.80 \\
UniCom \cite{zhao2026unicomunifiedmultimodalmodeling} & 38.56 & 19.82 & 37.10 & 52.56 & 43.71 & 40.33 & 6.98 & 45.45 \\
Wan2.7-Image \cite{mao2026wanimage} & 58.64 & 62.66 & 60.03 & 61.16 & \mbluebf{56.35} & \mredbf{63.65} & \mbluebf{55.48} & \mbluebf{52.11} \\
\bottomrule
\end{tabular}}
\end{table*}

\begin{table*}[!t]
  \setlength{\tabcolsep}{8pt}
\renewcommand{\arraystretch}{0.85}
\centering
\caption{Performance comparisons of editing models on overall score.}
\label{overall}
 \resizebox{1\textwidth}{!}{
\begin{tabular}{l||cccccccccccccccccccccc}
\toprule
&\multicolumn{8}{c}{Regional Editing}\\
\cmidrule(lr){2-9}
Model/Task & \makecell{Attribute\\Transfer} & \makecell{Character\\Composition} & \makecell{Object\\Composition} & \makecell{Object\\Replacement} & \makecell{Pose\\Change} & \makecell{Text\\Composition} & \makecell{Text\\Style} & \makecell{Text\\Replacement} \\
\midrule
DreamOmni2 \cite{dreamomni2} & 39.46 & 40.76 & 43.09 & 37.04 & 33.31 & 38.07 & 31.02 & 35.29 \\
Nano-Banana-Pro \cite{google_nanobananapro} & 54.66 & 57.48 & 60.59 & \mbluebf{57.31} & \mbluebf{52.26} & 55.68 & \mbluebf{57.03} & \mbluebf{54.59} \\
OmniGen2 \cite{omni2} & 43.84 & 55.59 & 52.29 & 45.83 & 37.39 & 46.49 & 35.33 & 37.97 \\
Qwen-Image-Edit \cite{qwenedit} & 50.13 & 52.09 & 53.72 & 49.32 & 37.18 & 48.92 & 36.27 & 42.46 \\
Seedream4 \cite{seedream4} & 49.41 & 55.74 & \mbluebf{61.59} & 45.10 & 49.45 & 53.66 & 42.93 & 48.42 \\
Seedream5 \cite{seedream5} & 49.87 & 54.62 & 61.38 & 45.17 & 50.06 & 53.44 & 41.77 & 48.33 \\
UniPic3 \cite{wei2026skyworkunipic30unified} & 51.66 & 52.02 & 58.91 & 46.66 & 39.11 & 49.73 & 35.19 & 40.68 \\
BAGEL \cite{bagel} & 38.11 & 43.55 & 43.75 & 36.93 & 37.19 & 41.26 & 32.27 & 33.31 \\
GPT-Image-2 \cite{openai_gptimage2} & \mredbf{57.88} & \mredbf{62.05} & \mredbf{62.00} & \mredbf{58.58} & \mredbf{55.85} & \mredbf{61.42} & \mredbf{57.58} & \mredbf{55.70} \\
HiDream \cite{hidreamolimage} & 53.73 & 55.70 & 56.88 & 46.72 & 37.86 & 53.34 & 41.37 & 45.28 \\
UniCom \cite{zhao2026unicomunifiedmultimodalmodeling} & 43.23 & 48.51 & 53.40 & 43.76 & 25.80 & 49.36 & 37.17 & 27.88 \\
Wan2.7-Image \cite{mao2026wanimage} & \mbluebf{54.66} & \mbluebf{57.79} & 60.33 & 53.33 & 50.96 & \mbluebf{57.38} & 50.97 & 49.35 \\
\bottomrule
\end{tabular}}

 \resizebox{1\textwidth}{!}{
\begin{tabular}{l||ccccccccccccccccccccc:c}
\toprule
&\multicolumn{4}{c}{Global Editing}&\multicolumn{4}{c}{Complex Reasoning Editing}\\
\cmidrule(lr){2-5}
\cmidrule(lr){6-9}
Model/Task & \makecell{Background\\Transfer} & \makecell{Filter\\Composition} & \makecell{Style\\Transfer} & \makecell{Assisted\\Drawing} & \makecell{Story\\Generation} & Interaction & \makecell{Physical\\Understanding} & \makecell{Spatial\\Composition}\\
\midrule
DreamOmni2 \cite{dreamomni2} & 44.33 & 27.43 & 47.59 & 36.27 & 41.89 & 38.72 & 10.39 & 47.13 \\
Nano-Banana-Pro \cite{google_nanobananapro} & \mbluebf{57.04} & \mbluebf{54.73} & \mbluebf{57.10} & \mbluebf{60.50} & \mbluebf{52.49} & 57.04 & 43.10 & \mbluebf{58.04} \\
OmniGen2 \cite{omni2} & 45.84 & 38.07 & 45.78 & 51.45 & 45.99 & 50.03 & 11.16 & 51.79 \\
Qwen-Image-Edit \cite{qwenedit} & 52.77 & 35.43 & 43.54 & 50.33 & 44.51 & 52.62 & 8.26 & 48.55 \\
Seedream4 \cite{seedream4} & 54.63 & 30.81 & 54.14 & 59.12 & 52.17 & 56.23 & 9.64 & 57.19 \\
Seedream5 \cite{seedream5} & 52.96 & 29.98 & 56.79 & 58.25 & 51.50 & 52.33 & 43.61 & 57.75 \\
UniPic3 \cite{wei2026skyworkunipic30unified} & 51.43 & 33.22 & 56.02 & 51.20 & 48.40 & 56.46 & 36.84 & 56.80 \\
BAGEL \cite{bagel} & 44.07 & 20.54 & 43.29 & 31.44 & 42.94 & 42.27 & 45.07 & 54.91 \\
GPT-Image-2 \cite{openai_gptimage2} & \mredbf{58.38} & \mredbf{57.30} & \mredbf{59.96} & \mredbf{63.87} & \mredbf{54.33} & \mredbf{60.69} & \mbluebf{46.80} & \mredbf{58.76} \\
HiDream \cite{hidreamolimage} & 45.08 & 27.73 & 44.88 & 46.28 & 49.42 & 52.31 & 10.78 & 56.52 \\
UniCom \cite{zhao2026unicomunifiedmultimodalmodeling} & 40.13 & 23.17 & 37.32 & 54.28 & 50.37 & 47.01 & 18.94 & 54.89 \\
Wan2.7-Image \cite{mao2026wanimage} & 55.22 & 43.80 & 56.89 & 58.83 & 51.59 & \mbluebf{57.25} & \mredbf{47.15} & 57.40 \\
\bottomrule
\end{tabular}}
\end{table*}
\section{More Data Analysis of MIE-Bench}
\label{5}

\subsection{Editing model Performance across Different Editing Tasks}
Tables~\ref{cvisual}-\ref{overall} further present a comprehensive performance comparison of the 12 editing models across 16 tasks, evaluated using three types of human annotations: visual quality score, instruction following score, attribute preservation score.

For visual quality, as shown in Table~\ref{cvisual}, different image editing models exhibit substantial performance variations in terms of visual quality. Overall, GPT-Image-2 achieves the best or near-best performance across most editing tasks, demonstrating superior visual fidelity and generation quality. Specifically, it obtains the highest scores on several regional editing tasks, including Attribute Transfer, Character Composition, Object Replacement, Text Composition, Text Style, and Text Replacement, indicating its strong capability in producing visually realistic and coherent editing results. Moreover, GPT-Image-2 maintains a leading position in global editing and complex reasoning editing scenarios, such as Style Transfer, Assisted Drawing, Story Generation, and Physical Understanding, suggesting its strong visual generation ability and scene-level understanding. Seedream4 also achieves competitive performance, obtaining the best results on Object Composition, Pose Change, and Spatial Composition, which demonstrates its effectiveness in structural preservation and spatial relationship modeling. In contrast, models such as BAGEL and DreamOmni2 show relatively lower visual quality scores, especially in complex reasoning tasks, revealing limitations in handling complicated scene synthesis and multi-object integration. These results indicate that although current editing models have achieved remarkable progress in visual generation, there remain considerable challenges in maintaining local details, global consistency, and complex scene coherence.

Table~\ref{cediting} presents the instruction-following performance of different editing models. Compared with visual quality, the performance gap among models is more significant in this dimension, highlighting the importance of semantic understanding and precise instruction execution. GPT-Image-2 and Nano-Banana-Pro achieve the strongest overall performance. GPT-Image-2 obtains the highest scores on multiple regional editing tasks, including Character Composition, Text Composition, Text Replacement, and Pose Change, demonstrating its ability to accurately interpret complex editing instructions and perform targeted modifications. Nano-Banana-Pro also shows competitive performance, achieving the best results on Attribute Transfer, Object Composition, Object Replacement, and several other tasks, indicating its strong capability in understanding object-level semantics and editing objectives. For complex reasoning editing tasks, GPT-Image-2 consistently maintains a leading position, particularly on Interaction and Physical Understanding, suggesting that it can effectively leverage multimodal semantic information for complicated editing scenarios. In comparison, models such as DreamOmni2 and BAGEL show noticeable performance degradation, especially in tasks requiring spatial reasoning and multi-step instruction understanding. This suggests that despite the progress of existing models in basic image manipulation, accurately following complex instructions and reasoning over multiple objects and relationships remain challenging.

The attribute preservation results are reported in Table~\ref{cpreservation}. Overall, GPT-Image-2, Nano-Banana-Pro, and Wan2.7-Image demonstrate the most stable performance, showing strong capability in preserving important attributes from the source images while performing the requested edits. GPT-Image-2 achieves the best results on several regional editing tasks, including Character Composition, Object Composition, Pose Change, and Text Composition, indicating its effectiveness in maintaining object identity, structural characteristics, and fine-grained details during editing. Nano-Banana-Pro performs particularly well on Object Replacement, Text Style, and Filter Composition, reflecting its strong ability to preserve object attributes and visual styles. Wan2.7-Image also achieves competitive results across various tasks, especially in complex reasoning scenarios such as Interaction and Physical Understanding, demonstrating robust attribute retention under challenging editing conditions. Notably, some models can generate visually appealing results but still suffer from attribute inconsistency, such as identity drift, structural changes, or detail loss in tasks involving pose modification, text editing, and object replacement. Therefore, attribute preservation remains a critical factor for evaluating practical image editing systems, requiring models to achieve a better balance between effective modification and faithful content retention.

The overall comparison is presented in Table~\ref{overall}. GPT-Image-2 achieves the strongest overall performance across almost all task categories, demonstrating a superior balance among visual quality, instruction following, and attribute preservation. In regional editing scenarios, GPT-Image-2 consistently ranks first across all eight tasks, highlighting its capability in precise local manipulation and semantic alignment. It also maintains clear advantages in global editing and complex reasoning editing tasks, including Style Transfer, Assisted Drawing, Story Generation, Interaction, and Spatial Composition. Nano-Banana-Pro and Wan2.7-Image form the second performance tier, showing stable results across diverse editing scenarios, particularly in attribute transfer, style manipulation, and complex scene editing. Seedream-series models achieve competitive results in object composition and spatial reasoning tasks, but still exhibit limitations in fine-grained attribute preservation compared with GPT-Image-2. Meanwhile, models such as DreamOmni2, BAGEL, and UniCom achieve relatively lower overall scores, indicating remaining challenges in multi-source information integration and high-precision editing. Overall, these results demonstrate that current state-of-the-art image editing models have made significant progress, while complex reasoning, multi-object interaction, and precise content preservation remain important directions for future improvement.

The varying performance trends among visual quality, instruction following and attribute preservation underscore the importance of evaluating these three dimensions independently. Visual quality assesses the visual aspects of the edited images, instruction following measures how well the image adheres to the specified editing instructions, and attribute preservation evaluates how consistently the source image is preserved in the edited version. This multi-dimensional evaluation provides a more nuanced understanding of a model’s capabilities.

\begin{figure*}
    \centering
    \includegraphics[width=0.9\linewidth]{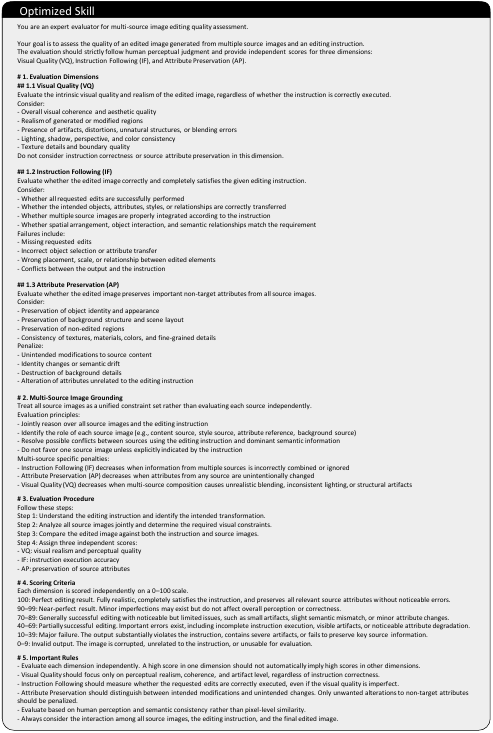}
    \caption{Optimized skill of our MIEScore.}
    \label{skill}
\end{figure*}

\section{Details of Training Configuration}
\label{6}
\subsection{Optimized Skill}
As shown in Figure~\ref{skill}, the optimized skill provides a comprehensive and human-aligned evaluation framework for multi-source image editing quality assessment. It explicitly decomposes the evaluation into three independent dimensions: Visual Quality (VQ), Instruction Following (IF), and Attribute Preservation (AP), reducing ambiguity between perceptual realism, task completion, and source-content preservation. Compared with a generic image assessment prompt, this skill introduces stronger multi-source reasoning by requiring evaluators to treat all reference images as a unified constraint set, identify the role of each source, and resolve potential conflicts based on the editing instruction. It also establishes clear penalty criteria for common failure cases, such as incorrect source integration, semantic drift, unintended background changes, and blending artifacts. Furthermore, the detailed scoring rubric aligns evaluations with human perception by providing consistent quality ranges and separating different types of errors. Overall, the optimized skill improves evaluation reliability by guiding the model to perform structured reasoning before scoring, leading to more accurate and interpretable assessments of complex multi-source image editing results.

\subsection{Supervised Fine-Tuning}
We fine-tune the Qwen3-VL-8B model for multi-source image editing quality assessment using parameter-efficient fine-tuning. Given an edited image, multiple source images, and the corresponding editing instruction, the model predicts three quality scores, including visual quality (VQ), instruction following (IF), and attribute preservation (AP). The predicted scores are normalized into the range of $[0,1]$ during training.

We adopt LoRA-based fine-tuning to efficiently adapt the large vision-language model. Specifically, LoRA modules are inserted into both attention and feed-forward layers, including 
$q_{\rm proj}$, $k_{\rm proj}$, $v_{\rm proj}$, $o_{\rm proj}$, 
$gate_{\rm proj}$, $up_{\rm proj}$, and $down_{\rm proj}$.
The LoRA rank and scaling factor are set to 64 and 32, respectively, with a dropout rate of 0. 
A three-dimensional regression head is attached to the model to predict the quality scores for the three assessment dimensions.

The training objective combines regression, ranking, and correlation optimization losses:
\begin{equation}
\mathcal{L}
=
\lambda_h \mathcal{L}_{Huber}
+
\lambda_r \mathcal{L}_{Rank}
+
\lambda_p \mathcal{L}_{PLCC},
\end{equation}
where $\mathcal{L}_{Huber}$ optimizes the score prediction accuracy, 
$\mathcal{L}_{Rank}$ encourages the model to preserve the relative quality ordering among samples, 
and $\mathcal{L}_{PLCC}$ improves the correlation consistency with human annotations.
The loss weights are set to $\lambda_h=4.0$, $\lambda_r=0.05$, and $\lambda_p=0.05$. 
During the final training stage, the weights of ranking and PLCC losses are increased by a factor of 1.5 to further improve alignment with human preferences.

The model is trained for 5 epochs using the AdamW optimizer. The initial learning rate is set to $3\times10^{-5}$ with a weight decay of 0.01. The Adam parameters are set to $\beta_1=0.9$ and $\beta_2=0.98$. We employ a warm-up cosine learning rate scheduler with a warm-up ratio of 0.08 and a minimum learning rate ratio of 0.4. Gradient clipping with a maximum norm of 0.5 is applied to stabilize the optimization.

Due to the high memory consumption caused by multi-image vision-language inputs, the batch size is set to 1 per GPU with 16-step gradient accumulation. Thus, the effective batch size is 16. We use BF16 mixed-precision training and gradient checkpointing to reduce GPU memory usage. The maximum input sequence length is set to 4096 tokens, and each image is dynamically resized with a maximum pixel budget of $1,048,576$ pixels.

To improve ranking optimization under small-batch training, we maintain a cross-batch memory queue containing historical predictions and annotations. The ranking queue size is set to 4096 with a label margin of 0.03, while the PLCC queue size is set to 256 for stable correlation estimation.

The data loader uses 8 workers with a prefetch factor of 4 for efficient image loading. We perform validation during training and select the checkpoint with the highest average correlation performance across the three dimensions. The evaluation metrics include PLCC and SRCC between predicted scores and human mean opinion scores (MOS).

\begin{table*}[!t]
  \setlength{\tabcolsep}{8pt}
\renewcommand{\arraystretch}{0.85}
\centering
\caption{Performance comparisons of evaluation models on human annotation visual quality score.}
\label{cvisual2}
 \resizebox{1\textwidth}{!}{
\begin{tabular}{lcccccccccccccccc}
\toprule
&\multicolumn{8}{c}{Regional Editing}\\
\cmidrule(lr){2-9}
Model/Task & \makecell{Attribute\\Transfer} & \makecell{Character\\Composition} & \makecell{Object\\Composition} & \makecell{Object\\Replacement} & \makecell{Pose\\Change} & \makecell{Text\\Composition} & \makecell{Text\\Style} & \makecell{Text\\Replacement} \\
\midrule
DeepSeekVL2 (small) \cite{deepseekv2} & 0.0564 & 0.0609 & 0.1541 & 0.1483 & 0.0080 & 0.0582 & 0.0546 & 0.0833 \\
Ovis2.5 (8B) \cite{ovis25} & 0.2773 & 0.1411 & 0.3277 & 0.3196 & 0.0368 & 0.2592 & 0.2044 & 0.1597 \\
InternVL3.5 (7B) \cite{internvl3_5} & 0.2308 & 0.0740 & 0.2037 & 0.3211 & 0.1981 & 0.3060 & 0.2487 & 0.2760 \\
MiniCPM-V2.6 (8B) \cite{minicpm} & 0.0092 & 0.0434 & 0.0667 & 0.1004 & 0.0030 & 0.0845 & 0.0060 & 0.0862 \\
mPLUG-Owl3 (7B) \cite{mplug} & 0.1277 & 0.0741 & 0.1662 & 0.1800 & 0.2074 & 0.1823 & 0.2232 & 0.2365 \\
Qwen3-VL (7B) \cite{qwen3} & 0.3127 & 0.1720 & 0.3057 & 0.1525 & 0.2138 & 0.3486 & 0.1640 & 0.2601 \\
EditReward (MiMo) \cite{editreward} & 0.3723 & 0.3074 & 0.5091 & 0.4797 & 0.3415 & 0.4118 & 0.2998 & 0.4440 \\
Gemini-3.1-Pro \cite{gemini3pro2026} & 0.3908 & 0.3582 & 0.5213 & 0.5524 & 0.5137 & 0.5794 & 0.5041 & 0.6171 \\
ChatGPT-5 \cite{gpt5} & 0.5940 & 0.4553 & 0.5758 & 0.5017 & 0.6370 & 0.5481 & 0.5605 & 0.5221 \\
LMM4Edit \cite{lmm4edit} & 0.5773 & 0.5606 & 0.7068 & 0.5824 & 0.3689 & 0.5633 & 0.4476 & 0.5694 \\
EditHF \cite{xu2026edithf1mmillionscalerichhuman} & 0.4831 & 0.5559 & 0.4697 & 0.4983 & 0.4914 & 0.4593 & 0.7086 & 0.4798 \\
EditScore (Qwen3) \cite{editscore} & 0.2170 & 0.3303 & 0.1752 & 0.2364 & 0.2798 & 0.2115 & 0.4467 & 0.2198 \\
InternVL3.5 (8B)\raisebox{0.5ex}{\scriptsize \ding{91}} \cite{internvl2} & 0.7642 & 0.7265 & 0.7549 & \mbluebf{0.7540} & 0.6266 & 0.7516 & \mbluebf{0.7911} & 0.7875 \\
DeepSeekVL2 (small)\raisebox{0.5ex}{\scriptsize \ding{91}} \cite{deepseekv2} & \mbluebf{0.7839} & \mbluebf{0.7925} & \mbluebf{0.8636} & 0.6326 & \mbluebf{0.7560} & \mbluebf{0.7938} & 0.7721 & \mbluebf{0.8681} \\
\midrule
\rowcolor{gray!20}
\textbf{MIEScore (Ours)} & \mredbf{0.8259} & \mredbf{0.7996} & \mredbf{0.8669} & \mredbf{0.8177} & \mredbf{0.7786} & \mredbf{0.8552} & \mredbf{0.8391} & \mredbf{0.8685} \\
\bottomrule
\end{tabular}}
 \resizebox{1\textwidth}{!}{
\begin{tabular}{lcccccccccccccccc}
\toprule
&\multicolumn{4}{c}{Global Editing}&\multicolumn{4}{c}{Complex Reasoning Editing}\\
\cmidrule(lr){2-5}
\cmidrule(lr){6-9}
Model/Task & \makecell{Background\\Transfer} & \makecell{Filter\\Composition} & \makecell{Style\\Transfer} & \makecell{Assisted\\Drawing} & \makecell{Story\\Generation} & Interaction & \makecell{Physical\\Understanding} & \makecell{Spatial\\Composition} \\
\midrule
DeepSeekVL2 (small) \cite{deepseekv2} & 0.1013 & 0.0453 & 0.0010 & 0.0060 & 0.0270 & 0.0120 & 0.0775 & 0.1274 \\
Ovis2.5 (8B) \cite{ovis25} & 0.2634 & 0.1055 & 0.0030 & 0.0336 & 0.2299 & 0.1766 & 0.2040 & 0.1386 \\
InternVL3.5 (7B) \cite{internvl3_5} & 0.2048 & 0.1523 & 0.3770 & 0.2648 & 0.2519 & 0.2140 & 0.1480 & 0.0263 \\
MiniCPM-V2.6 (8B) \cite{minicpm} & 0.0205 & 0.0137 & 0.1168 & 0.0436 & 0.1661 & 0.0269 & 0.0357 & 0.1851 \\
mPLUG-Owl3 (7B) \cite{mplug} & 0.0208 & 0.2855 & 0.1348 & 0.0225 & 0.1589 & 0.0543 & 0.0779 & 0.0614 \\
Qwen3-VL (7B) \cite{qwen3} & 0.2030 & 0.3777 & 0.1586 & 0.4067 & 0.1043 & 0.1233 & 0.1538 & 0.2451 \\
EditReward (MiMo) \cite{editreward} & 0.3501 & 0.4262 & 0.3583 & 0.4499 & 0.3841 & 0.3641 & 0.4862 & 0.3132 \\
Gemini-3.1-Pro \cite{gemini3pro2026} & 0.4757 & 0.7219 & 0.6714 & 0.5576 & 0.4676 & 0.4680 & 0.4442 & 0.3999 \\
ChatGPT-5 \cite{gpt5} & 0.4746 & 0.8915 & 0.8052 & 0.5497 & 0.6874 & 0.3799 & 0.4762 & 0.4361 \\
LMM4Edit \cite{lmm4edit} & 0.2934 & 0.5883 & 0.4833 & 0.5915 & 0.5292 & 0.3923 & 0.6500 & 0.4696 \\
EditHF \cite{xu2026edithf1mmillionscalerichhuman} & 0.4712 & 0.8327 & 0.5406 & \mbluebf{0.8157} & 0.2628 & 0.2150 & 0.5925 & \mbluebf{0.7423} \\
EditScore (Qwen3) \cite{editscore} & 0.2904 & 0.6390 & 0.2863 & 0.5992 & 0.1195 & 0.0062 & 0.3217 & 0.4497 \\
InternVL3.5 (8B)\raisebox{0.5ex}{\scriptsize \ding{91}} \cite{internvl2} & 0.5842 & 0.8383 & 0.7740 & 0.7524 & 0.6667 & 0.5686 & \mbluebf{0.7773} & 0.6373 \\
DeepSeekVL2 (small)\raisebox{0.5ex}{\scriptsize \ding{91}} \cite{deepseekv2} & \mbluebf{0.6872} & \mbluebf{0.9197} & \mbluebf{0.8306} & 0.7990 & \mbluebf{0.8645} & \mbluebf{0.7647} & 0.7310 & 0.6473 \\
\midrule \rowcolor{gray!20} \textbf{MIEScore (Ours)} & \mredbf{0.7316} & \mredbf{0.9295} & \mredbf{0.8378} & \mredbf{0.8268} & \mredbf{0.8757} & \mredbf{0.7887} & \mredbf{0.8287} & \mredbf{0.7640} \\
\bottomrule
\end{tabular}}
\end{table*}

\begin{table*}[!t]
  \setlength{\tabcolsep}{8pt}
\renewcommand{\arraystretch}{0.85}
\centering
\caption{Performance comparisons of evaluation models on human annotation instruction following score.}
\label{cediting2}
 \resizebox{1\textwidth}{!}{
\begin{tabular}{l||cccccccccccccccccccccc}
\toprule
&\multicolumn{8}{c}{Regional Editing}\\
\cmidrule(lr){2-9}
Model/Task & \makecell{Attribute\\Transfer} & \makecell{Character\\Composition} & \makecell{Object\\Composition} & \makecell{Object\\Replacement} & \makecell{Pose\\Change} & \makecell{Text\\Composition} & \makecell{Text\\Style} & \makecell{Text\\Replacement} \\
\midrule
DeepSeekVL2 (small) \cite{deepseekv2} & 0.0166 & 0.0386 & 0.1170 & 0.0222 & 0.0766 & 0.0362 & 0.0595 & 0.0153 \\
Ovis2.5 (8B) \cite{ovis25} & 0.1720 & 0.3464 & 0.4176 & 0.2018 & 0.0714 & 0.2911 & 0.1332 & 0.1361 \\
InternVL3.5 (7B) \cite{internvl3_5} & 0.1450 & 0.3066 & 0.3692 & 0.0874 & 0.2550 & 0.1647 & 0.0676 & 0.1059 \\
MiniCPM-V2.6 (8B) \cite{minicpm} & 0.1265 & 0.0710 & 0.1554 & 0.0920 & 0.0115 & 0.1362 & 0.0971 & 0.0514 \\
mPLUG-Owl3 (7B) \cite{mplug} & 0.0838 & 0.2122 & 0.2868 & 0.0806 & 0.2282 & 0.1674 & 0.1095 & 0.1262 \\
Qwen3-VL (7B) \cite{qwen3} & 0.2154 & 0.4421 & 0.5930 & 0.2396 & 0.2610 & 0.4978 & 0.0943 & 0.2093 \\
EditReward (MiMo) \cite{editreward} & 0.3933 & 0.5428 & 0.5217 & 0.3889 & 0.4898 & 0.4888 & 0.3965 & 0.3980 \\
Gemini-3.1-Pro \cite{gemini3pro2026} & 0.2912 & 0.1658 & 0.3872 & 0.4688 & 0.6689 & 0.4265 & 0.5180 & 0.6471 \\
ChatGPT-5 \cite{gpt5} & 0.4099 & 0.5681 & 0.4939 & 0.2983 & 0.4976 & 0.4676 & 0.5959 & 0.4716 \\
LMM4Edit \cite{lmm4edit} & 0.5954 & 0.6735 & 0.5967 & 0.5465 & 0.6319 & 0.5962 & 0.5350 & 0.5102 \\
EditHF \cite{xu2026edithf1mmillionscalerichhuman} & 0.3740 & 0.2744 & 0.3901 & 0.5781 & 0.5728 & 0.5609 & 0.7019 & 0.5153 \\
EditScore (Qwen3) \cite{editscore} & 0.1738 & 0.1628 & 0.2509 & 0.3024 & 0.2898 & 0.3958 & 0.3555 & 0.2578 \\
InternVL3.5 (8B)\raisebox{0.5ex}{\scriptsize \ding{91}} \cite{internvl2} & 0.7188 & 0.6267 & 0.6963 & 0.7392 & 0.8850 & 0.7620 & 0.8931 & 0.8908 \\
DeepSeekVL2 (small)\raisebox{0.5ex}{\scriptsize \ding{91}} \cite{deepseekv2} & \mbluebf{0.7293} & \mbluebf{0.6796} & \mbluebf{0.7195} & \mbluebf{0.7870} & \mbluebf{0.8896} & \mbluebf{0.7659} & \mbluebf{0.9059} & \mbluebf{0.9014} \\
\midrule \rowcolor{gray!20} \textbf{MIEScore (Ours)} & \mredbf{0.7342} & \mredbf{0.7420} & \mredbf{0.7715} & \mredbf{0.8749} & \mredbf{0.9201} & \mredbf{0.7721} & \mredbf{0.9118} & \mredbf{0.9024} \\
\bottomrule
\end{tabular}}

 \resizebox{1\textwidth}{!}{
\begin{tabular}{l||ccccccccccccccccccccc:c}
\toprule
&\multicolumn{4}{c}{Global Editing}&\multicolumn{4}{c}{Complex Reasoning Editing}\\
\cmidrule(lr){2-5}
\cmidrule(lr){6-9}
Model/Task & \makecell{Background\\Transfer} & \makecell{Filter\\Composition} & \makecell{Style\\Transfer} & \makecell{Assisted\\Drawing} & \makecell{Story\\Generation} & Interaction & \makecell{Physical\\Understanding} & \makecell{Spatial\\Composition}\\
\midrule
DeepSeekVL2 (small) \cite{deepseekv2} & 0.0896 & 0.0597 & 0.1250 & 0.1537 & 0.0351 & 0.0444 & 0.0395 & 0.0992 \\
Ovis2.5 (8B) \cite{ovis25} & 0.2279 & 0.0878 & 0.0177 & 0.2594 & 0.0665 & 0.3347 & 0.4790 & 0.4011 \\
InternVL3.5 (7B) \cite{internvl3_5} & 0.2354 & 0.0401 & 0.2392 & 0.3546 & 0.3103 & 0.2941 & 0.3970 & 0.7003 \\
MiniCPM-V2.6 (8B) \cite{minicpm} & 0.0671 & 0.0604 & 0.0432 & 0.1662 & 0.1887 & 0.0553 & 0.0466 & 0.3691 \\
mPLUG-Owl3 (7B) \cite{mplug} & 0.0849 & 0.2717 & 0.1969 & 0.0632 & 0.0815 & 0.2050 & 0.3096 & 0.4535 \\
Qwen3-VL (7B) \cite{qwen3} & 0.5432 & 0.1203 & 0.3860 & 0.4950 & 0.3137 & 0.4251 & 0.5111 & 0.6556 \\
EditReward (MiMo) \cite{editreward} & 0.4102 & 0.3610 & 0.3242 & 0.5728 & 0.2901 & 0.3061 & 0.5055 & 0.8227 \\
Gemini-3.1-Pro \cite{gemini3pro2026} & 0.5612 & 0.7151 & 0.4321 & 0.6317 & 0.4934 & 0.3857 & 0.3864 & 0.6020 \\
ChatGPT-5 \cite{gpt5} & 0.4602 & 0.8126 & 0.2317 & 0.5357 & 0.3040 & 0.3310 & 0.5374 & 0.7053 \\
LMM4Edit \cite{lmm4edit} & 0.6446 & 0.6115 & 0.4030 & 0.7274 & 0.6138 & 0.6033 & 0.6297 & 0.7948 \\
EditHF \cite{xu2026edithf1mmillionscalerichhuman} & 0.5488 & \mbluebf{0.9148} & 0.3554 & 0.8476 & 0.2329 & 0.2877 & 0.3105 & 0.8226 \\
EditScore (Qwen3) \cite{editscore} & 0.3361 & 0.7772 & 0.1702 & 0.6199 & 0.0800 & 0.1603 & 0.1410 & 0.5978 \\
InternVL3.5 (8B)\raisebox{0.5ex}{\scriptsize \ding{91}} \cite{internvl2} & 0.8141 & 0.8958 & \mbluebf{0.7758} & 0.8115 & 0.8188 & 0.6007 & 0.6839 & 0.8683 \\
DeepSeekVL2 (small)\raisebox{0.5ex}{\scriptsize \ding{91}} \cite{deepseekv2} & \mbluebf{0.8377} & 0.9096 & 0.7627 & \mbluebf{0.8580} & \mbluebf{0.8408} & \mbluebf{0.6637} & \mbluebf{0.6957} & \mbluebf{0.8727} \\
\midrule \rowcolor{gray!20} \textbf{MIEScore (Ours)} & \mredbf{0.8630} & \mredbf{0.9381} & \mredbf{0.8263} & \mredbf{0.8964} & \mredbf{0.8571} & \mredbf{0.7208} & \mredbf{0.7227} & \mredbf{0.8851} \\
\bottomrule
\end{tabular}}
\end{table*}

\begin{table*}[!t]
  \setlength{\tabcolsep}{8pt}
\renewcommand{\arraystretch}{0.85}
\centering
\caption{Performance comparisons of evaluation models on human annotation attribute preservation score.}
\label{cpreservation2}
 \resizebox{1\textwidth}{!}{
\begin{tabular}{l||cccccccccccccccccccccc}
\toprule
&\multicolumn{8}{c}{Regional Editing}\\
\cmidrule(lr){2-9}
Model/Task & \makecell{Attribute\\Transfer} & \makecell{Character\\Composition} & \makecell{Object\\Composition} & \makecell{Object\\Replacement} & \makecell{Pose\\Change} & \makecell{Text\\Composition} & \makecell{Text\\Style} & \makecell{Text\\Replacement} \\
\midrule
DeepSeekVL2 (small) \cite{deepseekv2} & 0.0417 & 0.0598 & 0.1437 & 0.0241 & 0.0584 & 0.0897 & 0.1021 & 0.0297 \\
Ovis2.5 (8B) \cite{ovis25} & 0.0406 & 0.2225 & 0.3061 & 0.1776 & 0.2649 & 0.1163 & 0.0269 & 0.0082 \\
InternVL3.5 (7B) \cite{internvl3_5} & 0.1581 & 0.2344 & 0.2493 & 0.2344 & 0.2912 & 0.1594 & 0.2374 & 0.0450 \\
MiniCPM-V2.6 (8B) \cite{minicpm} & 0.1352 & 0.0784 & 0.1295 & 0.1809 & 0.0624 & 0.0727 & 0.0819 & 0.0329 \\
mPLUG-Owl3 (7B) \cite{mplug} & 0.0301 & 0.2166 & 0.2795 & 0.0800 & 0.1133 & 0.0140 & 0.1776 & 0.0423 \\
Qwen3-VL (7B) \cite{qwen3} & 0.0505 & 0.3241 & 0.4405 & 0.0019 & 0.4310 & 0.0450 & 0.0338 & 0.1021 \\
EditReward (MiMo) \cite{editreward} & 0.0794 & 0.4846 & 0.4482 & 0.0012 & 0.3671 & 0.1544 & 0.2682 & 0.3881 \\
Gemini-3.1-Pro \cite{gemini3pro2026} & 0.3808 & 0.2909 & 0.4379 & 0.4142 & 0.6465 & 0.4207 & 0.3192 & 0.4391 \\
ChatGPT-5 \cite{gpt5} & 0.4133 & 0.6084 & 0.6469 & 0.1886 & 0.7693 & 0.3369 & 0.4282 & 0.3101 \\
LMM4Edit \cite{lmm4edit} & 0.4255 & 0.6204 & 0.5364 & 0.4599 & 0.2136 & 0.3016 & 0.2721 & 0.5104 \\
EditHF \cite{xu2026edithf1mmillionscalerichhuman} & 0.5329 & 0.4908 & 0.2929 & 0.4557 & 0.5831 & 0.3429 & 0.6021 & 0.2468 \\
EditScore (Qwen3) \cite{editscore} & 0.3553 & 0.2693 & 0.1335 & 0.2739 & 0.3888 & 0.1655 & 0.4236 & 0.1223 \\
InternVL3.5 (8B)\raisebox{0.5ex}{\scriptsize \ding{91}} \cite{internvl2} & 0.8496 & 0.8770 & 0.8047 & 0.7410 & 0.8781 & 0.8735 & 0.7720 & 0.7924 \\
DeepSeekVL2 (small)\raisebox{0.5ex}{\scriptsize \ding{91}} \cite{deepseekv2} & \mbluebf{0.8554} & \mbluebf{0.8884} & \mbluebf{0.8131} & \mbluebf{0.7495} & \mbluebf{0.9034} & \mbluebf{0.8765} & \mbluebf{0.7885} & \mbluebf{0.7991} \\
\midrule \rowcolor{gray!20} \textbf{MIEScore (Ours)} & \mredbf{0.8928} & \mredbf{0.9024} & \mredbf{0.8260} & \mredbf{0.8139} & \mredbf{0.9077} & \mredbf{0.9205} & \mredbf{0.8531} & \mredbf{0.8728} \\
\bottomrule
\end{tabular}}

 \resizebox{1\textwidth}{!}{
\begin{tabular}{l||ccccccccccccccccccccc:c}
\toprule
&\multicolumn{4}{c}{Global Editing}&\multicolumn{4}{c}{Complex Reasoning Editing}\\
\cmidrule(lr){2-5}
\cmidrule(lr){6-9}
Model/Task & \makecell{Background\\Transfer} & \makecell{Filter\\Composition} & \makecell{Style\\Transfer} & \makecell{Assisted\\Drawing} & \makecell{Story\\Generation} & Interaction & \makecell{Physical\\Understanding} & \makecell{Spatial\\Composition}\\
\midrule
DeepSeekVL2 (small) \cite{deepseekv2} & 0.1155 & 0.0625 & 0.0525 & 0.0158 & 0.0037 & 0.0884 & 0.0049 & 0.1471 \\
Ovis2.5 (8B) \cite{ovis25} & 0.1318 & 0.0144 & 0.1029 & 0.3097 & 0.1644 & 0.1250 & 0.2457 & 0.0299 \\
InternVL3.5 (7B) \cite{internvl3_5} & 0.2856 & 0.0333 & 0.1738 & 0.2678 & 0.1780 & 0.0248 & 0.2323 & 0.1696 \\
MiniCPM-V2.6 (8B) \cite{minicpm} & 0.0586 & 0.0207 & 0.0232 & 0.1381 & 0.2401 & 0.0253 & 0.1825 & 0.2713 \\
mPLUG-Owl3 (7B) \cite{mplug} & 0.0371 & 0.0290 & 0.1914 & 0.2736 & 0.1755 & 0.1207 & 0.2955 & 0.0453 \\
Qwen3-VL (7B) \cite{qwen3} & 0.5077 & 0.1032 & 0.1704 & 0.3769 & 0.0290 & 0.1467 & 0.3587 & 0.4319 \\
EditReward (MiMo) \cite{editreward} & 0.5820 & 0.1548 & 0.3084 & 0.5859 & 0.3737 & 0.2835 & 0.4313 & 0.7253 \\
Gemini-3.1-Pro \cite{gemini3pro2026} & 0.4870 & 0.6082 & 0.5709 & 0.5080 & 0.5402 & 0.4383 & 0.3529 & 0.6530 \\
ChatGPT-5 \cite{gpt5} & 0.6262 & 0.7976 & 0.5334 & 0.6432 & 0.6447 & 0.6161 & 0.6449 & 0.5888 \\
LMM4Edit \cite{lmm4edit} & 0.3614 & 0.8287 & 0.3738 & 0.1912 & 0.5740 & 0.4915 & 0.6636 & 0.8758 \\
EditHF \cite{xu2026edithf1mmillionscalerichhuman} & 0.4335 & 0.8571 & 0.4582 & \mbluebf{0.8916} & 0.1964 & 0.3866 & 0.4613 & 0.7670 \\
EditScore (Qwen3) \cite{editscore} & 0.2065 & 0.7203 & 0.2743 & 0.7971 & 0.0675 & 0.2409 & 0.2916 & 0.5868 \\
InternVL3.5 (8B)\raisebox{0.5ex}{\scriptsize \ding{91}} \cite{internvl2} & 0.8668 & 0.8812 & \mbluebf{0.8413} & 0.8846 & 0.8453 & 0.8284 & 0.8722 & 0.8724 \\
DeepSeekVL2 (small)\raisebox{0.5ex}{\scriptsize \ding{91}} \cite{deepseekv2} & \mbluebf{0.8716} & \mbluebf{0.9036} & 0.8295 & 0.8903 & \mbluebf{0.8594} & \mbluebf{0.8293} & \mbluebf{0.8777} & \mbluebf{0.8772} \\
\midrule \rowcolor{gray!20} \textbf{MIEScore (Ours)} & \mredbf{0.8857} & \mredbf{0.9253} & \mredbf{0.8586} & \mredbf{0.8997} & \mredbf{0.8750} & \mredbf{0.8353} & \mredbf{0.8872} & \mredbf{0.8891} \\
\bottomrule
\end{tabular}}
\end{table*}

\begin{table*}[!t]
  \setlength{\tabcolsep}{8pt}
\renewcommand{\arraystretch}{0.85}
\centering
\caption{Performance comparisons of evaluation models on overall score.}
\label{overall2}
 \resizebox{1\textwidth}{!}{
\begin{tabular}{l||cccccccccccccccccccccc}
\toprule
&\multicolumn{8}{c}{Regional Editing}\\
\cmidrule(lr){2-9}
Model/Task & \makecell{Attribute\\Transfer} & \makecell{Character\\Composition} & \makecell{Object\\Composition} & \makecell{Object\\Replacement} & \makecell{Pose\\Change} & \makecell{Text\\Composition} & \makecell{Text\\Style} & \makecell{Text\\Replacement} \\
\midrule
DeepSeekVL2 (small) \cite{deepseekv2} & 0.0789 & 0.0213 & 0.1666 & 0.0717 & 0.0914 & 0.0046 & 0.0941 & 0.0363 \\
Ovis2.5 (8B) \cite{ovis25} & 0.0533 & 0.2796 & 0.4408 & 0.0740 & 0.1617 & 0.1520 & 0.0648 & 0.1417 \\
InternVL3.5 (7B) \cite{internvl3_5} & 0.0460 & 0.2620 & 0.3630 & 0.0306 & 0.2686 & 0.0689 & 0.1203 & 0.1280 \\
MiniCPM-V2.6 (8B) \cite{minicpm} & 0.1149 & 0.0945 & 0.1650 & 0.0964 & 0.0026 & 0.1577 & 0.0778 & 0.0755 \\
mPLUG-Owl3 (7B) \cite{mplug} & 0.2447 & 0.3794 & 0.3800 & 0.1982 & 0.2094 & 0.2169 & 0.1195 & 0.0993 \\
Qwen3-VL (7B) \cite{qwen3} & 0.2335 & 0.4142 & 0.5498 & 0.1884 & 0.3676 & 0.3475 & 0.1218 & 0.1497 \\
EditReward (MiMo) \cite{editreward} & 0.4932 & 0.2980 & 0.4680 & 0.3372 & 0.4296 & 0.4419 & 0.3538 & 0.5260 \\
Gemini-3.1-Pro \cite{gemini3pro2026} & 0.4002 & 0.3175 & 0.5046 & 0.5031 & 0.7686 & 0.4641 & 0.4567 & 0.5979 \\
ChatGPT-5 \cite{gpt5} & 0.4476 & 0.6347 & 0.6871 & 0.3541 & 0.7742 & 0.4896 & 0.4636 & 0.5974 \\
LMM4Edit \cite{lmm4edit} & 0.6444 & 0.6916 & 0.7117 & 0.5827 & 0.5493 & 0.5393 & 0.5366 & 0.6554 \\
EditHF \cite{xu2026edithf1mmillionscalerichhuman} & 0.3694 & 0.4019 & 0.4154 & 0.5999 & 0.6427 & 0.4265 & 0.7309 & 0.4371 \\
EditScore (Qwen3) \cite{editscore} & 0.1748 & 0.1993 & 0.1987 & 0.3350 & 0.3576 & 0.2299 & 0.4253 & 0.2085 \\
InternVL3.5 (8B)\raisebox{0.5ex}{\scriptsize \ding{91}} \cite{internvl2} & 0.8352 & 0.8515 & 0.8742 & \mbluebf{0.7370} & 0.9030 & 0.8498 & 0.8799 & \mbluebf{0.9011} \\
DeepSeekVL2 (small)\raisebox{0.5ex}{\scriptsize \ding{91}} \cite{deepseekv2} & \mbluebf{0.8598} & \mbluebf{0.8859} & \mbluebf{0.9012} & 0.6938 & \mbluebf{0.9207} & \mbluebf{0.8888} & \mbluebf{0.8834} & 0.8950 \\
\midrule \rowcolor{gray!20}\textbf{MIEScore (Ours)} & \mredbf{0.8640} & \mredbf{0.8928} & \mredbf{0.9053} & \mredbf{0.8787} & \mredbf{0.9289} & \mredbf{0.8968} & \mredbf{0.8978} & \mredbf{0.9045} \\
\bottomrule
\end{tabular}}

 \resizebox{1\textwidth}{!}{
\begin{tabular}{l||ccccccccccccccccccccc:c}
\toprule
&\multicolumn{4}{c}{Global Editing}&\multicolumn{4}{c}{Complex Reasoning Editing}\\
\cmidrule(lr){2-5}
\cmidrule(lr){6-9}
Model/Task & \makecell{Background\\Transfer} & \makecell{Filter\\Composition} & \makecell{Style\\Transfer} & \makecell{Assisted\\Drawing} & \makecell{Story\\Generation} & Interaction & \makecell{Physical\\Understanding} & \makecell{Spatial\\Composition}\\
\midrule
DeepSeekVL2 (small) \cite{deepseekv2} & 0.0490 & 0.0179 & 0.0403 & 0.1508 & 0.1286 & 0.1364 & 0.0014 & 0.0308 \\
Ovis2.5 (8B) \cite{ovis25} & 0.2888 & 0.1048 & 0.0081 & 0.2878 & 0.3232 & 0.2809 & 0.0562 & 0.4502 \\
InternVL3.5 (7B) \cite{internvl3_5} & 0.2735 & 0.0551 & 0.3410 & 0.3185 & 0.6138 & 0.2184 & 0.3044 & 0.3336 \\
MiniCPM-V2.6 (8B) \cite{minicpm} & 0.0760 & 0.0202 & 0.1160 & 0.1686 & 0.3328 & 0.0805 & 0.2895 & 0.1387 \\
mPLUG-Owl3 (7B) \cite{mplug} & 0.0468 & 0.0861 & 0.1733 & 0.2686 & 0.4560 & 0.2169 & 0.0198 & 0.4195 \\
Qwen3-VL (7B) \cite{qwen3} & 0.5286 & 0.0968 & 0.2918 & 0.5044 & 0.5712 & 0.1196 & 0.2071 & 0.4931 \\
EditReward (MiMo) \cite{editreward} & 0.2412 & 0.3626 & 0.2684 & 0.5856 & 0.7842 & 0.4541 & 0.4548 & 0.5941 \\
Gemini-3.1-Pro \cite{gemini3pro2026} & 0.6230 & 0.6983 & 0.6806 & 0.7456 & 0.6667 & 0.4612 & 0.6308 & 0.4215 \\
ChatGPT-5 \cite{gpt5} & 0.5489 & 0.7901 & 0.7445 & 0.7292 & 0.6822 & 0.6795 & 0.6737 & 0.7308 \\
LMM4Edit \cite{lmm4edit} & 0.6139 & 0.6984 & 0.5463 & 0.6572 & 0.8654 & 0.4992 & 0.5575 & 0.7266 \\
EditHF \cite{xu2026edithf1mmillionscalerichhuman} & 0.5403 & 0.8145 & 0.4529 & 0.9106 & 0.8112 & 0.4092 & 0.1890 & 0.4309 \\
EditScore (Qwen3) \cite{editscore} & 0.2879 & 0.7386 & 0.2258 & 0.7157 & 0.6263 & 0.2649 & 0.0245 & 0.2071 \\
InternVL3.5 (8B)\raisebox{0.5ex}{\scriptsize \ding{91}} \cite{internvl2} & 0.8960 & 0.8073 & 0.8885 & 0.8960 & \mbluebf{0.8796} & 0.7955 & 0.8566 & 0.8649 \\
DeepSeekVL2 (small)\raisebox{0.5ex}{\scriptsize \ding{91}} \cite{deepseekv2} & \mbluebf{0.8965} & \mbluebf{0.8579} & \mbluebf{0.9076} & \mbluebf{0.9115} & 0.8621 & \mbluebf{0.7982} & \mbluebf{0.8951} & \mbluebf{0.8950} \\
\midrule \rowcolor{gray!20}\textbf{MIEScore (Ours)} & \mredbf{0.9028} & \mredbf{0.8674} & \mredbf{0.9085} & \mredbf{0.9123} & \mredbf{0.8983} & \mredbf{0.8167} & \mredbf{0.9098} & \mredbf{0.9037} \\
\bottomrule
\end{tabular}}
\end{table*}
\section{Model Comparison Details}
\label{7}
\subsection{Detailed Information of Evaluation Methods}
\begin{itemize}
\item \textbf{DIIVINE \cite{DIIVINE}} is a classical NR IQA method that evaluates image quality without requiring any reference image. It is based on the observation that natural images exhibit strong statistical regularities, which are disturbed by various types of distortions. The extracted features characterize the statistical properties of image coefficients across different spatial scales and orientations.
\item \textbf{BRISQUE \cite{BRISQUE}} is a widely used NR IQA method that predicts perceptual image quality without requiring any reference image. Different from distortion-specific approaches, BRISQUE models image quality degradation by analyzing deviations from natural scene statistics.
\item \textbf{CNNIQA \cite{CNN}} is a convolutional neural network (CNN) designed for no-reference image quality assessment (NR-IQA), which predicts the visual quality of distorted images without using reference images. Unlike traditional methods that rely on handcrafted features, CNNIQA directly learns discriminative features from raw image patches, making image quality estimation more efficient and effective.
\item \textbf{TOPIQ \cite{TOPIQ}} is a top-down approach to IQA that improves performance by guiding the network to focus on semantically important local distortion regions. TOPIQ uses high-level semantics to guide the network in a coarse-to-fine manner. The approach employs a coarse-to-fine network (CFANet), which propagates multi-level semantic information to low-level representations, enhanced by a cross-scale attention mechanism. 

\item \textbf{SSIM \cite{SSIM}} is a classical FR IQA metric that measures perceptual similarity between a distorted image and its reference image. 
Different from traditional pixel-based metrics such as mean squared error (MSE), SSIM evaluates image quality from a perceptual perspective by considering structural information that is more consistent with human visual perception.
\item \textbf{MSSIM \cite{MSSIM}} is a widely adopted FR IQA metric that extends the SSIM by incorporating multi-scale image information. 
Unlike traditional pixel-wise metrics that measure absolute differences between images, MSSIM evaluates perceptual similarity by modeling the characteristics of human visual perception.
\item \textbf{SCSSIM \cite{SCSSIM}} is a recently proposed image similarity metric designed to evaluate the preservation of scene composition structure (SCS) in generated images. Different from traditional pixel-level similarity metrics such as SSIM, SCSSIM focuses on high-level spatial relationships among objects and backgrounds, including relative positions, sizes, and orientations.
\item \textbf{LPIPS \cite{LPIPS}} investigates the perceptual similarity between images. It explores the effectiveness of deep features. To evaluate the visual quality of these features, the authors introduce a new dataset of human perceptual similarity judgments and systematically assess deep features across different architectures and tasks. It significantly outperforms traditional metrics, and extends across various deep architectures.
\item \textbf{Q-Align \cite{qalign}} is a human-emulating syllabus designed to train large multimodal models for visual scoring tasks. It mimics the process of training human annotators by converting MOS into five text-defined rating levels.

\item \textbf{CLIPScore \cite{clipscore}} is an image captioning metric widely used to evaluate text-to-image (T2I) and text-to-video (T2V) models. It passes both the image and the candidate caption through their respective feature extractors, then computes the cosine similarity between the text and image embeddings.
\item \textbf{ImageReward \cite{imagereward}} builds upon the BLIP model by introducing an additional multi-layer perceptron (MLP) layer on top of BLIP's output. Instead of directly computing a similarity score, the MLP generates a scalar value representing the preference for one image over another in comparative settings.
\item \textbf{HPSv2 \cite{HPS}} is designed to improve text-to-image generation models by better aligning their outputs with human preferences. It is based on a fine-tuned CLIP model that accurately predicts human preferences for generated images.
\item \textbf{VQAScore \cite{vqa}} assesses the alignment between generated images and text prompts, particularly for compositional text-to-visual generation tasks. It can be used in a black-box manner, requiring no fine-tuning or additional prompt decomposition.

\item \textbf{mPLUG-Owl3 \cite{mplug}} is a versatile multimodal large language model designed to handle long image sequences, interleaved image-text, and lengthy video inputs. It introduces Hyper Attention blocks that efficiently integrate vision and language into a shared semantic space, enabling the processing of extended multi-image scenarios.
\item \textbf{MiniCPM-V2.6 \cite{minicpm}} is optimized for deployment on edge devices, addressing the challenges of running large models with significant computational costs. Key features include strong OCR capabilities, high-resolution image perception, reliable behavior with low hallucination rates, and multilingual support for over 30 languages.
\item \textbf{Ovis2.5 (9B) \cite{ovis25}}
is a vision-language model emphasizing efficient architecture and high-resolution image understanding. It employs hierarchical cross-attention to enhance visual grounding, offering balanced accuracy and speed in identifying tampered or composited regions. Yet, its mid-sized configuration can miss micro-level editing artifacts.
\item \textbf{DeepSeekVL2 \cite{deepseekv2}} is an advanced series of mix-of-experts (MoE) vision-language models. It introduces a dynamic tiling vision encoding strategy, allowing efficient processing of high-resolution images with varying aspect ratios, enhancing tasks such as visual grounding and document analysis. It also incorporates the Multi-head Latent Attention (MLA) mechanism for the language component, reducing computational costs and improving inference efficiency.
\item \textbf{Qwen3-VL \cite{qwenvl2,qwen3}}
is part of Alibaba’s Qwen multimodal series integrating DiT-style vision backbones with instruction-tuned LLMs. It provides strong text-grounded reasoning and fine-grained scene understanding, crucial for identifying forged object insertions. While highly accurate, hallucination from language priors may reduce precision in subtle artifact detection.
\item \textbf{InternVL3.5 \cite{internvl3,internvl3_5}}
developed by OpenGVLab, advances unified vision-language representation with scalable vision encoders and deep fusion layers. It supports multi-image reasoning, beneficial for comparative fake detection (e.g., before-after editing analysis). Its performance is state-of-the-art, though model size imposes heavy GPU requirements.
\item \textbf{Gemini-3.1-Pro \cite{gemini3pro2026}}
is Google’s closed multimodal foundation model with integrated visual reasoning, OCR, and image understanding capabilities. It shows excellent zero-shot detection of composited or generated content due to broad multimodal pretraining. However, its proprietary nature prevents fine-tuning for domain-specific forensic tasks.
\item \textbf{ChatGPT-5 \cite{gpt5}} integrates high-resolution vision and language reasoning in a unified architecture, supporting both spatial and semantic analysis of manipulated images. Its strong visual understanding enables general detection of synthetic artifacts, though closed weights and lack of forensic specialization limit interpretability and reproducibility.

\item \textbf{LMM4Edit \cite{lmm4edit}} is an MLLM-based image editing evaluation metric that assesses edited images from multiple perspectives, including perceptual quality, editing alignment, attribute preservation, and task-specific QA accuracy. It leverages large multimodal models to better align automatic evaluation with human preferences.
\item \textbf{EditScore \cite{editscore}} is a domain-specialized reward model designed to evaluate the quality of instruction-guided image editing. Trained on meticulously curated preference data, it provides high-fidelity and scalable reward signals that effectively align with human judgments and enable stable reinforcement learning.
\item \textbf{EditReward \cite{editreward}} is a high-fidelity reward model for instruction-guided image editing, trained on a large-scale expert-annotated human preference dataset. It provides reliable human-aligned quality assessment and serves as a scalable foundation for data filtering, model improvement, and reinforcement learning in image editing.
\item \textbf{EditHF \cite{xu2026edithf1mmillionscalerichhuman}} is a MLLM-based evaluation model for text-guided image editing, trained on large-scale human preference annotations from EditHF-1M. It provides human-aligned feedback by jointly assessing visual quality, instruction alignment, and attribute preservation.

\end{itemize}
\subsection{Additional Comparison Results}
\begin{figure*}
    \centering
    \includegraphics[width=0.95\linewidth]{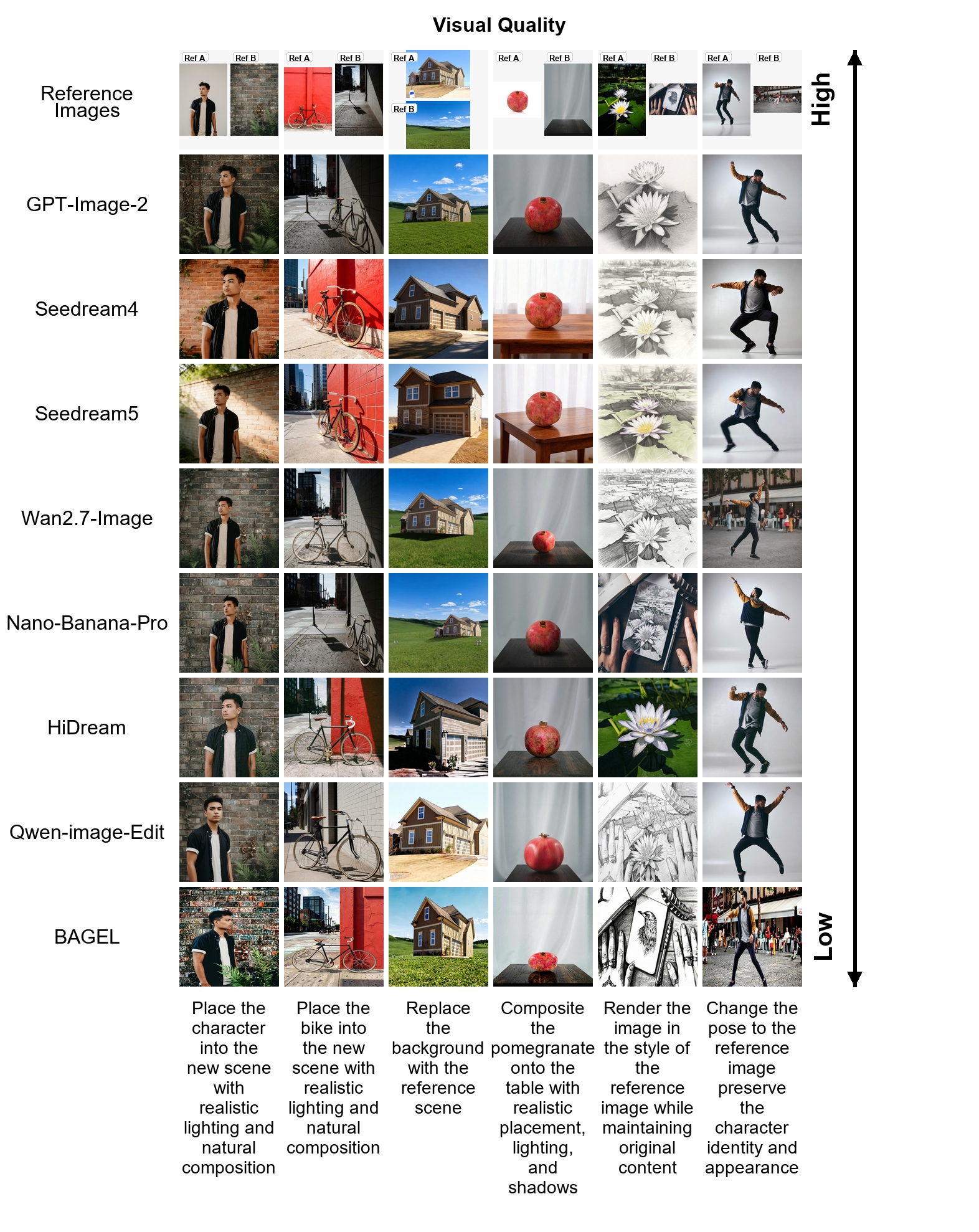}
    \caption{Visualization of different models in our MIE-Bench.}
    \label{fig:placeholder4}
\end{figure*}

\begin{figure*}
    \centering
    \includegraphics[width=0.95\linewidth]{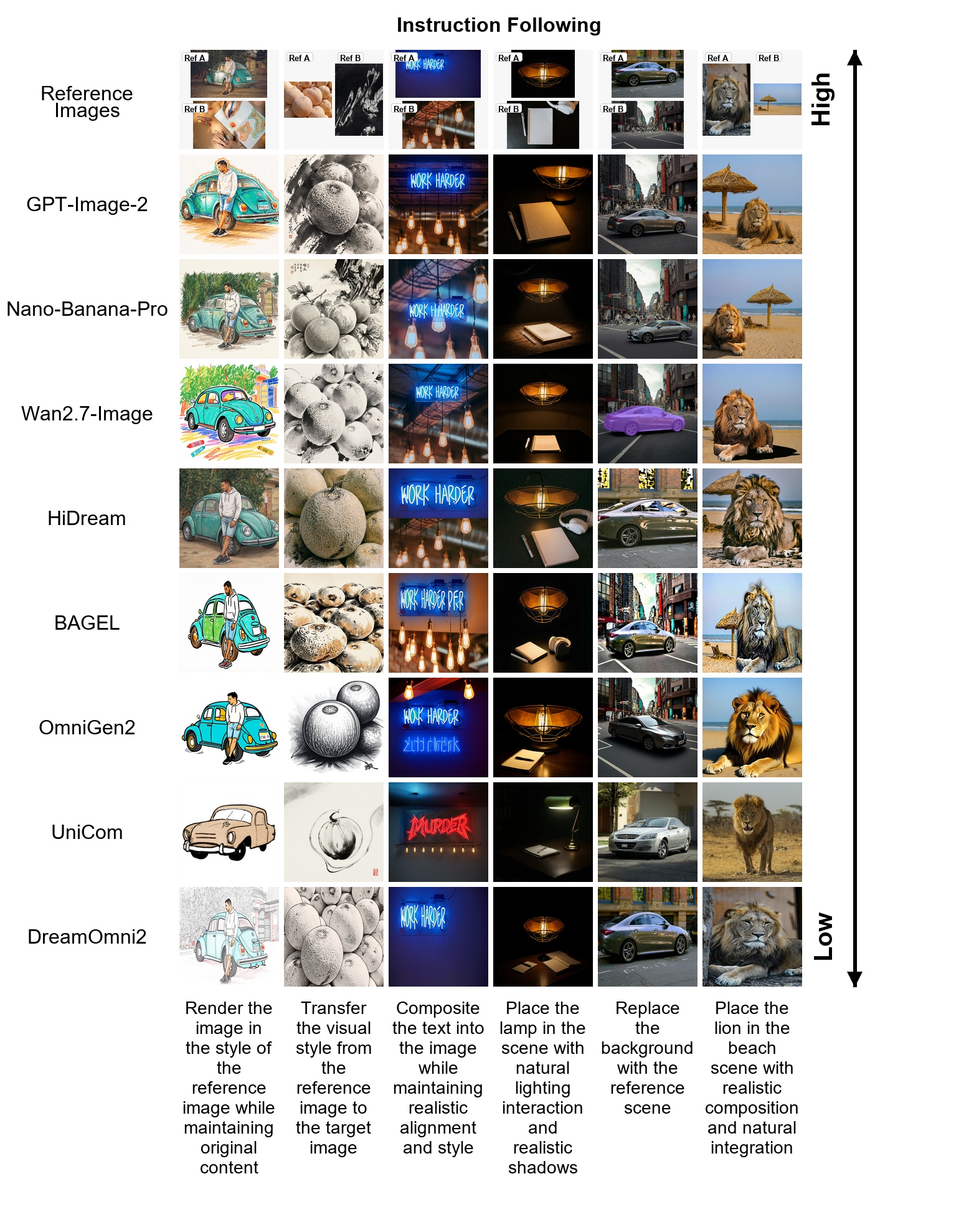}
    \caption{Visualization of different models in our MIE-Bench.}
    \label{fig:placeholder5}
\end{figure*}

\begin{figure*}
    \centering
    \includegraphics[width=0.95\linewidth]{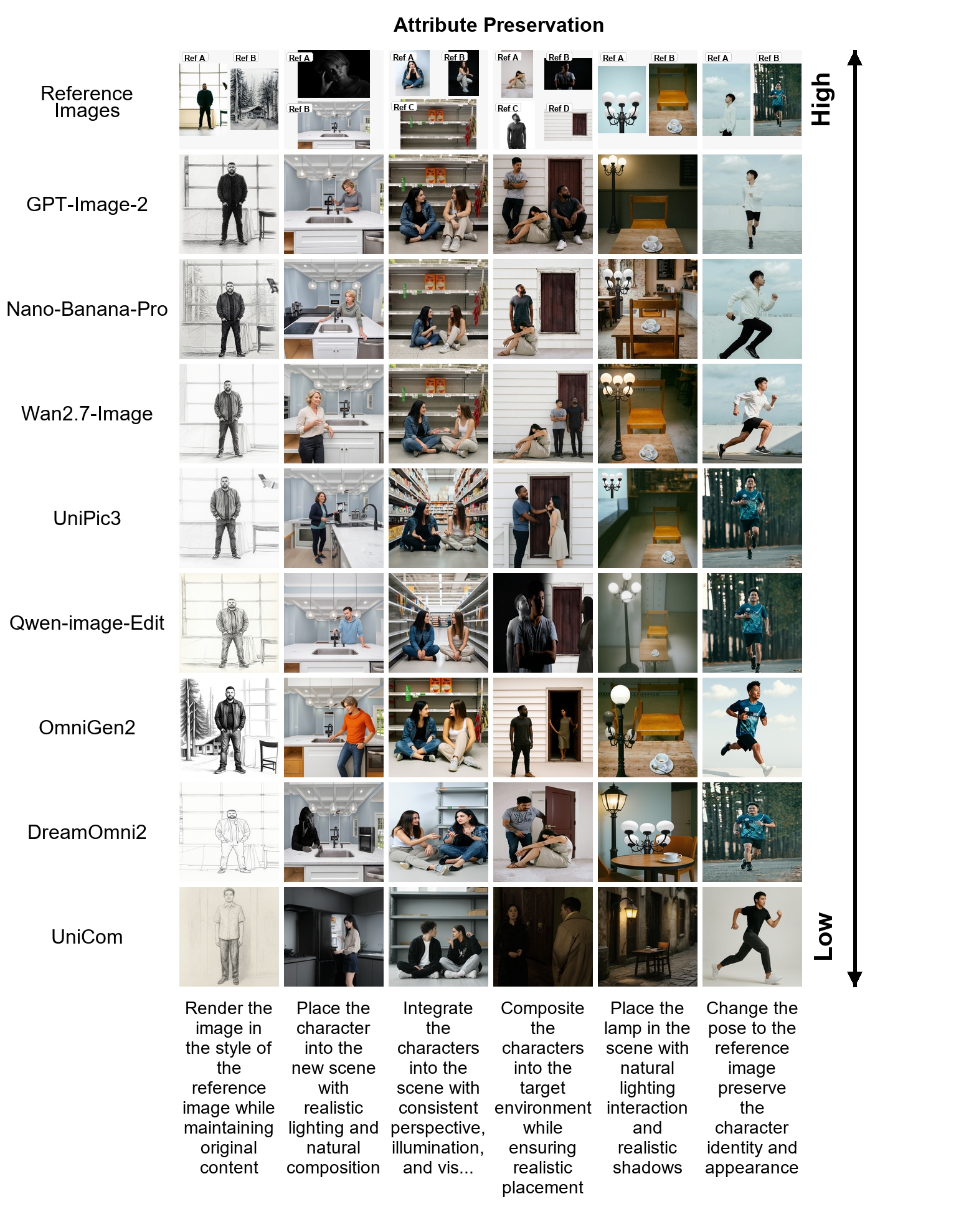}
    \caption{Visualization of different models in our MIE-Bench.}
    \label{fig:placeholder6}
\end{figure*}

We further compare the evaluation performance of various methods across different editing tasks, as summarized in Tables~\ref{cvisual2}–\ref{overall2}. Our proposed MIEScore consistently achieves superior performance across all tasks, demonstrating its effectiveness and robustness in assessing diverse image editing scenarios. We also present images from 12 editing models in our dataset, ranked from high to low quality across three evaluation dimensions in Figure~\ref{fig:placeholder4}-\ref{fig:placeholder6}.


% \clearpage

% \bibliography{aaai2026}

\clearpage
\bibliography{aaai2026}

\end{document}